%% file: main_author_info.tex
\documentclass{article}

\usepackage{preamble} 

\title{Interpreting Learning Under Competing Models: Joint and Stepwise Approaches for Dynamic Cognitive Diagnosis}

\author[1]{Yawen Ma}
\author[2]{Sahoko Ishida}
\author[3]{Kate Cain}
\author[1]{Gabriel Wallin\thanks{Corresponding author. School of Mathematical
Sciences, Lancaster University, Lancaster, LA1 4YF, United Kingdom.
\texttt{g.wallin@lancaster.ac.uk}}}
\affil[1]{School of Mathematical Sciences, Lancaster University, Lancaster,
LA1 4YF, Lancashire, United Kingdom}
\affil[2]{Department of Computer Science, University of Oxford, Oxford,
United Kingdom}
\affil[3]{Department of Psychology, Lancaster University, Lancaster,
LA1 4YF, Lancashire, United Kingdom}
\date{}

\begin{document}

\maketitle

\abstract{Digital learning environments record learners’ responses to individual items, making it possible to study the development of specific skills rather than overall scores. Drawing conclusions about learning from these data requires a model that links responses to latent skills and tracks how mastery changes over time. When the skills measured by each item are unknown, the analyst must decide whether to estimate this structure, the Q-matrix, jointly with the learning process, or to establish it first and study learning afterwards. We show that this decision can change substantive conclusions about how learners develop. Using dynamic cognitive diagnostic models, we analyse data from two reading games measuring vocabulary and comprehension from Grade 2 to Grade 3, with item-text embeddings providing prior information for the unknown Q-matrix. A joint analysis and a bias-corrected stepwise analysis agree that most learners move toward mastering both skills, but disagree about how many remain only partially proficient at Grade 3, changing how reading progress would be reported. A simulation study identifies when the two analyses diverge and shows that joint analysis is more reliable when the item-skill structure is uncertain and the item pool changes between grades. We provide R code for both analyses.}

\keywords{Cognitive Diagnostic Models; Dynamic Cognitive Diagnosis; $Q$-matrix Estimation; Bayesian Inference; Learning Transitions;
Natural Language Processing.}

\section{Introduction}
A single test score says little about which skills a learner has and has not mastered. Learning involves changes in how component skills are acquired and coordinated over time, and a total score cannot reveal them. Reading is a good example. Skilled reading depends on more than accurate and fluent word recognition; it also requires listening comprehension
\citep{larrc2015learning, gough1986decoding}, which itself is based on vocabulary, sentence processing, and the identification of ideas within prose \citep{larrc2017pressure, oakhill2012precursors}. Digital learning environments make it possible to observe these component skills at the level of individual
items. As learners interact with an educational tool, the tool produces log files that record not only whether each item was answered correctly, but also timestamps, actions, response times, repeated attempts, and early exit behaviour. These records allow researchers to move beyond aggregated performance and to study how a learner's knowledge develops through repeated interaction with the measurement instrument.

This richness creates an interpretive problem: how should these records be translated into statements about latent skills and their development? Because learners are observed repeatedly as they progress, the data describe a mastery profile that changes over time rather than a fixed one, and how that change is described depends on the statistical model. The model does more than determine estimation accuracy; it shapes the substantive account of what was learned, by whom, and when. Cognitive diagnostic models (CDMs) are well suited to this task, because they classify learners by mastery or non-mastery of several latent
attributes rather than placing them on a single scale \citep{haertel1984application, junker2001cognitive, templin2010diagnostic}.
Dynamic extensions of CDMs allow mastery to change across occasions, which makes them appropriate for longitudinal digital data \citep{wang2018tracking, wang2020development, zhan2018cognitive, zhan2019using, liang2023latent}. Applying them to real data, however, forces several decisions: how to use the response data, how to specify or estimate the item-attribute structure (the $Q$-matrix),
how to incorporate external information, and whether learning transitions are estimated jointly with the measurement model or in separate steps.

This article treats the last of these decisions as a substantive question rather than a technical one. Using data from two digital reading games, we ask whether estimating the measurement model and the learning process jointly, or in separate steps, leads to different conclusions about learners' mastery, their transitions between mastery states, and the variables associated with learning. A joint analysis estimates the measurement model, the $Q$-matrix, the mastery trajectories, and the transition model together. A bias-corrected stepwise analysis estimates the measurement model first and then fits the transition
model from the resulting classifications, with a correction for classification error. We find that the two analyses agree on the broad direction of learning but differ in how confidently learners are assigned to full rather than partial mastery, and that this difference is largest when the measurement structure is uncertain, as in our data.

Because the games were not designed around a predefined item-attribute structure, the attributes each item measures are not known in advance, and the $Q$-matrix must be estimated. We use item text to inform this estimation:
sentence embeddings of the item content provide prior information about the item-attribute structure, which the response data can then revise. Text-informed priors of this kind are a recent addition to cognitive diagnosis rather than a standard component. \citet{ma2026nlp} introduced them within a jointly estimated
CDM. Implementing them within a bias-corrected stepwise procedure, as we do here, has not to our knowledge been done before, and it is what allows the comparison to be fair: the joint and stepwise analyses then draw on the same
item text, response data, and covariates, so any difference between them reflects the estimation strategy and not the information available to each model.

The remainder of the article is organised as follows. We first review the use of response data for modelling learning, including dynamic extensions and the role of the $Q$-matrix, and describe how item text can inform the measurement structure when the $Q$-matrix is unknown. We then discuss how log-derived
covariates and learner characteristics support interpretation, and set out the joint and stepwise strategies. We apply the framework to the reading-game data and examine how the modelling choice affects conclusions about learning, and we use simulations to establish when the two strategies recover the $Q$-matrix, the mastery profiles, and the transition parameters. The article closes with the
implications and limitations of the approach.

\subsection{Using Response Data to Model Learning Processes}
Response data have long been the primary source of evidence in educational and psychological measurement. In traditional applications, item responses are often summarized by total scores or modeled using item response theory to estimate a general proficiency level. Such approaches are useful when the goal is to locate students on a common ability scale, but they provide limited information about the specific skills that students have or have not mastered. This limitation is important in learning environments where the goal is not only to evaluate performance but also to provide diagnostic information that can guide feedback, instruction, and intervention.

CDMs address this problem by linking observed responses to a vector of latent attribute mastery indicators \citep{haertel1984application, junker2001cognitive, templin2010diagnostic}. Instead of representing students using a single continuous trait, CDMs classify students into mastery profiles that describe their status on multiple skills. This makes CDMs especially useful for educational applications where interpretable information about component skills is needed. For reading data, such a framework is appealing because reading development involves multiple related but distinct skills, and students may show different patterns of mastery across decoding, fluency, vocabulary, and other comprehension-related attributes.

When data are collected repeatedly over time, static CDMs are insufficient because they do not directly model changes in mastery. Dynamic CDMs and related longitudinal variants extend the CDM framework by allowing latent attribute profiles to evolve across time \citep{wang2018tracking, wang2020development, zhan2018cognitive, zhan2019using, zhan2020partial, liang2023latent}. These models provide a way to study learning transitions rather than only cross-sectional mastery status. In digital learning environments, this dynamic perspective is crucial because students interact with tasks repeatedly and may acquire skills during the period of observation. However, the interpretation of such learning transitions depends on the measurement structure, the available external information, and the estimation strategy used to connect observed responses to latent mastery states.

\subsection{Using Item Text to Inform the Measurement}
A central component of CDMs is the $Q$-matrix, which specifies which latent attributes are required by each item. The $Q$-matrix determines the meaning of the latent attributes and directly affects item parameter estimation, mastery profiles, and conclusions about student learning. Misspecification of the $Q$-matrix can therefore lead to biased inferences about both items and learners \citep{rupp2008effects, chen2015statistical}. This issue becomes even more consequential in dynamic CDMs, because uncertainty in the measurement structure can influence the interpretation of learning transitions.

In many real-world digital learning environments, the $Q$-matrix is not fully known, and different domain experts may provide inconsistent specifications. As a result, the relation between item content and latent diagnostic attributes may remain uncertain. A growing body of research has therefore treated the $Q$-matrix as an object of inference rather than as a fixed input. Bayesian and data-driven approaches have been proposed to estimate, validate, or revise the $Q$-matrix using response data \citep{chen2018bayesian, culpepper2016revisiting, gu2021sufficient, fang2019identifiability}. These developments are important because they recognize that measurement structure is itself uncertain in many applied settings.

At the same time, response data alone may not always identify the item-attribute structure with sufficient certainty. This problem is especially likely in short assessments, sparse adaptive trajectories, or settings in which items are heterogeneous in content. Digital learning environments often provide additional information that can help reduce this uncertainty. In particular, item text and response options contain semantic information about what an item asks students to do. Recent developments in natural language processing make it possible to represent such text using embedding-based methods that capture semantic relationships among words, sentences, or item components \citep{vaswani2017attention, devlin2019bert, reimers2019sentencebert}. For CDMs, text-derived information need not determine the $Q$-matrix directly. Instead, text-derived information serves as structured prior information, allowing item content to inform the estimation of plausible item-attribute relations. 

Among embedding-based approaches, Sentence-BERT (SBERT) is particularly useful for representing assessment text because it produces sentence-level embeddings that can be efficiently compared using similarity measures such as cosine similarity or Euclidean distance \citep{reimers2019sentencebert}. SBERT extends the BERT network to generate semantically meaningful vector representations, allowing semantically similar texts to be located close to one another in a shared embedding space. Compared with standard BERT representations, SBERT substantially improves computational efficiency for large-scale semantic similarity and clustering tasks while maintaining strong performance on semantic textual similarity benchmarks \citep{reimers2019sentencebert}. In the context of educational assessment, these properties make SBERT especially suitable for quantifying semantic relationships among items and response options. Such semantic similarities may provide useful information about the complexity and potential attribute requirements of items, which can then be incorporated into the prior structure of the $Q$-matrix estimation process. Recent work by \citep{ma2026nlp} used SBERT-derived information to construct informative priors for $Q$-matrix estimation and demonstrated its effectiveness on a different dataset.

\subsection{Incorporating External Information for Interpreting Learning Processes}

Digital learning data contain more than item responses. They may also include log-based summaries, response times, number of attempts, success frequencies, early exit behaviors, learner demographics, and prior achievement. These sources of information are useful because they describe aspects of the learning process that are not fully captured by binary correctness. For example, two students may produce the same response pattern but differ substantially in time spent, persistence, number of attempts, or prior literacy ability. Such differences may affect how learning trajectories should be interpreted.

The present study analysed log files to understand student learning on two games in a supplementary digital reading support. In addition, an out-of-game assessment, the Dynamic Indicators of Basic Early Literacy Skills \citep{universityoforegon2018dibels}, was administered to measure each student's initial literacy ability. Students were classified into four performance levels: well below benchmark, below benchmark, at benchmark, and above benchmark. These categories were used to place students at an appropriate starting level in the app. Additional categorical covariates included race, special educational needs (SEN), English language learner status (ELL), and gender. Continuous log-based covariates included the average number of attempts, the number of correctly answered questions, and average response time. Descriptive statistics for the continuous and categorical covariates are summarized in Tables \ref{tab:summary_continuous} and \ref{tab:summary_categorical}, respectively.

Including such information can improve the substantive interpretation of learning trajectories by connecting latent changes to observable features of students and their learning environments. However, incorporating external information also creates methodological challenges. If covariates are measured with error, unevenly observed, or strongly related to the adaptive item selection process, then their inclusion may affect both estimation and interpretation. For this reason, external information should be treated not merely as additional predictors, but as part of an interpretive framework. In applied educational settings, researchers often want to know not only whether students learned, but under what conditions learning occurred and whether different groups of students followed different developmental patterns. Dynamic CDMs provide a structured way to address these questions, but the conclusions depend on how response data, item information, and external covariates are combined.

\subsection{Joint and Stepwise Strategies for Dynamic CDMs}

A final modeling decision concerns whether the measurement and structural components of a dynamic CDM should be estimated jointly or in separate steps. Stepwise approaches have a long history in latent class, latent transition, latent profile, growth mixture, and related finite mixture models. In these approaches, researchers first estimate a measurement model and then relate estimated latent classes or profiles to covariates, transitions, or distal outcomes \citep{Bakk2013, ClarkMuthen2009, Asparouhov2014, Bakk2018, Bolck2004, Vermunt2010}. The main advantage of stepwise methods is flexibility. They allow researchers to establish a measurement model before adding structural components and to modify covariate or transition models without refitting the full model.

However, simple stepwise procedures may treat assigned latent classes as if they were observed without error. This can bias structural parameter estimates when classification uncertainty is ignored \citep{Croon2002}. Bias-adjusted three-step methods address this problem by incorporating information about classification error into subsequent structural analyses \citep{Asparouhov2014, Bakk2013, Bakk2018, Bolck2004, Vermunt2010}. These developments are highly relevant for dynamic CDMs because learning trajectories are usually inferred from uncertain latent mastery classifications rather than directly observed.

Joint approaches represent a different strategy. They estimate measurement parameters, latent mastery profiles, transition parameters, and covariate effects within a single model. By accounting for uncertainty across model components, joint estimation may provide more stable inference when tests are short, sample sizes are modest, latent states are difficult to distinguish, or the $Q$-matrix is uncertain \citep{ma2026dynamic}. This feature is especially relevant for transition parameters, which are often difficult to estimate accurately when transition models are fitted after latent profile classification \citep{liang2023latent}. Related one-step approaches have been used in latent variable modeling and cognitive diagnosis to estimate measurement and structural components simultaneously \citep{DelaTorre2004, Ayers2013, Park2014, wang2019joint}. However, joint estimation can also be less convenient when researchers wish to modify parts of the structural model or compare alternative measurement specifications.

The distinction between joint and stepwise strategies is therefore not only computational, but also interpretive. A stepwise analysis asks how learning conclusions change after a measurement model has first been established and then linked to transitions and external variables. A joint analysis asks which learning trajectories are most coherent with the measurement model, transition model, covariates, and observed responses simultaneously. These two inferential pathways can agree when the measurement model is strong and classification uncertainty is limited, but they may diverge when the $Q$-matrix is unknown, item information is sparse, or latent states are difficult to distinguish. This article studies that difference as an applied measurement problem.

\section{Methods}

\subsection{Cognitive Diagnostic Measurement Model}

Cognitive diagnostic models (CDMs) are widely used in educational, psychological, and behavioral measurement to analyze item response data and provide fine-grained classifications of individuals' mastery or non-mastery of discrete latent attributes \citep{huff2007diagnostic, zenisky2012developing, roussos2007cognitive}. In this paper, an attribute refers to a specific skill or ability required for answering items correctly. CDMs provide a framework for identifying which attributes are likely to have been mastered by each individual and examining which attributes are required by each item. In this framework, the relationship between items and attributes is commonly represented by a $Q$-matrix, whose entries indicate whether a given item requires a given attribute \citep{tatsuoka1983rule, de2009dina}. Different CDMs make different assumptions about how required attributes combine to produce a correct response. Non-compensatory models assume that all required attributes must be mastered, whereas compensatory models allow mastery of some required attributes to compensate for non-mastery of others. More general models, such as the generalized DINA model, relax the assumptions of the deterministic inputs, noisy ``and'' gate (DINA) and deterministic inputs, noisy ``or'' gate (DINO) models by allowing the effects of required attributes to vary across items \citep{de2011generalized}.

In this paper, the response of student $i$ to item $j$ at time $t$ is denoted by $Y_{ijt} \in \{0,1\}$, where $i=1,\ldots,N$ denotes students, $t=1,\ldots,T$ denotes time points, $j=1,\ldots,J_t$ denotes items administered at time $t$, and $k=1,\ldots,K$ denotes latent attributes. The latent mastery profile of student $i$ at time $t$ is
$$
\boldsymbol{\alpha}_{it}
=
(\alpha_{i1t},\ldots,\alpha_{iKt})^{\top},
\qquad
\alpha_{ikt}\in\{0,1\},
$$

where $\alpha_{ikt}=1$ indicates that student $i$ has mastered attribute $k$ at time $t$. The item-attribute relationship at time $t$ is represented by a $J_t \times K$ $Q$-matrix,

$$
Q_t
=
(q_{jkt}),
\qquad
q_{jkt}\in\{0,1\}.
$$

The entry $q_{jkt}=1$ indicates that item $j$ at time $t$ requires attribute $k$.

A commonly used non-compensatory CDM is the DINA model \citep{junker2001cognitive, de2009dina}. Under the DINA model, the ideal response indicator is

\begin{align}
\eta_{ijt}
&=
\prod_{k=1}^{K}
\alpha_{ikt}^{q_{jkt}} .
\end{align}

Thus, $\eta_{ijt}=1$ only when student $i$ has mastered all attributes required by item $j$ at time $t$. Let $g_{jt}$ denote the guessing parameter and $s_{jt}$ denote the slipping parameter. The guessing parameter represents the probability of a correct response when the ideal response is incorrect, and the slipping parameter represents the probability of an incorrect response when the ideal response is correct. The item response probability is

\begin{align}
\label{eq:responseprobability}
P(Y_{ijt}=1 \mid \boldsymbol{\alpha}_{it},Q_t,g_{jt},s_{jt})
&=
(1-s_{jt})^{\eta_{ijt}}g_{jt}^{1-\eta_{ijt}} . 
\end{align}

Although the proposed Bayesian framework can be implemented with different CDM measurement components, the empirical and simulation analyses in this paper use the DINA model. This choice is made for three reasons. First, the DINA model has a simple non-compensatory structure, which is straightforward to interpret when an item is assumed to require mastery of all specified attributes. Second, its guessing and slipping parameters provide a computationally efficient measurement component for joint estimation with an unknown $Q$-matrix and dynamic latent mastery trajectories. Third, using the DINA model maintains consistency with previous applications of this modeling framework \citep{ma2026dynamic}. For simplicity, however, without loss of generality, we present the proposed approach under the DINA measurement model\footnote{An R package has been developed to allow switching between different measurement models and will be made available upon acceptance of the manuscript.}. 

Under the DINA measurement component, the likelihood contribution at time $t$ is

\begin{align}
p(Y_t \mid \boldsymbol{\alpha}_t,Q_t,\mathbf g_t,\mathbf s_t)
&=
\prod_{i=1}^{N}
\prod_{j=1}^{J_t}
\left\{
(1-s_{jt})^{\eta_{ijt}}g_{jt}^{1-\eta_{ijt}}
\right\}^{Y_{ijt}}
\left\{
s_{jt}^{\eta_{ijt}}(1-g_{jt})^{1-\eta_{ijt}}
\right\}^{1-Y_{ijt}} .
\end{align}

\subsection{Unknown \texorpdfstring{$Q$}{Q}-matrix and Text-informed Prior}

In practice, such as in educational technology applications, the $Q$-matrix is not known with certainty. An expert-specified item-attribute structure may not always be available, and different experts may disagree about which attributes are required for a given item. To address this uncertainty, we propose estimating $Q_t$ from the response data while incorporating item-level text information as prior information.

Let $\tau_{jt}$ denote a standardized text-derived signal for item $j$ at time $t$. The $\tau_{jt}$ is standardized before entering the prior for $Q_t$. Let $\pi_{jkt}\in(0,1)$ denote the prior probability that item $j$ at time $t$ requires attribute $k$, that is, the prior inclusion probability that $q_{jkt}=1$. We model this probability as:

\begin{align}
\mathrm{logit}(\pi_{jkt})
&=
\mathrm{logit}(\theta)
-
\lambda \tau_{jt}, \\
q_{jkt}\mid \theta,\lambda,\tau_{jt}
&\sim
\mathrm{Bernoulli}(\pi_{jkt}) .
\end{align}

Here, $\theta \in (0,1)$ controls the overall sparsity of the $Q$-matrix, and $\lambda \in \mathbb{R}$ controls how strongly the text-derived signal modifies this prior inclusion probability. When $\lambda=0$, the prior reduces to a Bernoulli prior in which each $q_{jkt}$ equals one with probability $\theta$. The negative sign reflects the working assumption that items with stronger text-based specificity tend to require fewer attributes, although the posterior distribution can override this prior when the response data provide contrary evidence.

In this application, the text-derived signal is used as auxiliary information about the likely complexity of each item. Because the available text information is summarised at the item level, the prior modifies the overall probability that an item requires an attribute but does not, by itself, determine which attribute is required. Attribute-specific $Q$-matrix structure is therefore informed by the combination of response data, admissibility constraints, and the dynamic model. This choice reflects the information available in the empirical application and is intended to provide a text-informed prior rather than a standalone NLP-based $Q$-matrix estimation procedure.

To ensure identifiability, the $Q$-matrices are restricted to satisfy three conditions. First, each item must require at least one attribute, so zero rows in the $Q$-matrix are not allowed. Second, each attribute must be measured by at least three items at each time point. Third, each attribute must have at least one single-attribute item, meaning that for every attribute there is at least one item that requires that attribute and no other attributes \citep{gu2021sufficient}.

The prior distributions are specified using weakly informative ranges that reflect the scale of each parameter. The baseline inclusion probability satisfies $\theta\in(0,1)$ and is assigned a beta prior,
\begin{align}
\theta
&\sim
\mathrm{Beta}(a_{\theta},b_{\theta}).
\end{align}
We specified a $\mathrm{Beta}(6,4)$ prior for $\theta$, which has a mean of $0.6$ and a variance of approximately $0.0218$.

The text effect $\lambda$ is a real-valued coefficient and is assigned a normal prior,

\begin{align}
\lambda
&\sim
N(0,\sigma_{\lambda}^{2}).
\end{align}
In the empirical analyses, we set $\sigma_{\lambda}=0.5$, so that $\lambda\sim N(0,0.5^2)$.

The DINA item parameters satisfy $g_{jt}\in(0,1)$ and $s_{jt}\in(0,1)$. They are assigned beta priors, in line with previous works \citep{ma2026dynamic},

\begin{align}
g_{jt}
&\sim
\mathrm{Beta}(1,1), \\
s_{jt}
&\sim
\mathrm{Beta}(1,1).
\end{align}

\subsection{Dynamic Structural Model}

The structural component describes students' initial mastery and subsequent learning transitions over time. Let $Z_{i0}$ denote covariates measured before the first assessment occasion, and let $Z_{i,t-1}$ denote covariates available before the transition from time $t-1$ to time $t$. These covariates may include student-level background variables, initial language ability measures, or time-varying learning process variables.

For the initial time point, mastery of attribute $k$ is modeled using a logistic regression:
\begin{align}
\mathrm{logit}\{P(\alpha_{ik1}=1\mid Z_{i0})\}
&=
\beta_{0k}
+
Z_{i0}^{\top}\boldsymbol{\beta}_{k}.
\end{align}
Here, $\beta_{0k}$ denotes the intercept for initial mastery of attribute $k$, and $\boldsymbol{\beta}_{k}$ denotes the corresponding regression coefficients.

For transitions between time points, the main parameter of interest is the acquisition probability from non-mastery to mastery. For $t=2,\ldots,T$, this probability is modeled as
\begin{align}
\mathrm{logit}\{P(\alpha_{ikt}=1\mid \alpha_{ik,t-1}=0,Z_{i,t-1})\}
&=
\gamma_{01,k,0}
+
Z_{i,t-1}^{\top}\boldsymbol{\gamma}_{01,k}.
\end{align}

Here, $\gamma_{01,k,0}$ denotes the intercept for acquiring attribute $k$, and $\boldsymbol{\gamma}_{01,k}$ denotes the effect of covariates on the acquisition probability. When loss of mastery is allowed, the transition from mastery to non-mastery is modeled as
\begin{align}
\mathrm{logit}\{P(\alpha_{ikt}=0\mid \alpha_{ik,t-1}=1,Z_{i,t-1})\}
&=
\gamma_{10,k,0}
+
Z_{i,t-1}^{\top}\boldsymbol{\gamma}_{10,k}.
\end{align}

The empirical and simulation analyses focus primarily on $\boldsymbol{\gamma}_{01,k}$ because these parameters directly describe learning, which describes the probability of acquiring an attribute among students who had not previously mastered it. The loss parameters $\boldsymbol{\gamma}_{10,k}$ are included when the application allows mastery status to decline over time.

The acquisition and loss probabilities for one attribute and two time points can be written as
\begin{align}
P(\alpha_{ik2}=1\mid \alpha_{ik1}=0,Z_{i1})
&=
\frac{
\exp(\gamma_{01,k,0}+Z_{i1}^{\top}\boldsymbol{\gamma}_{01,k})
}{
1+\exp(\gamma_{01,k,0}+Z_{i1}^{\top}\boldsymbol{\gamma}_{01,k})
}, \\
P(\alpha_{ik2}=0\mid \alpha_{ik1}=1,Z_{i1})
&=
\frac{
\exp(\gamma_{10,k,0}+Z_{i1}^{\top}\boldsymbol{\gamma}_{10,k})
}{
1+\exp(\gamma_{10,k,0}+Z_{i1}^{\top}\boldsymbol{\gamma}_{10,k})
}.
\end{align}

The regression parameters are assigned normal priors. Continuous covariates were standardized before model fitting. In the joint empirical analysis, the initial mastery parameters and transition parameters were assigned as:

\begin{align}
\beta_{0k},\boldsymbol{\beta}_{k}
&\sim
N(0,1), \\
\gamma_{01,k,0},\boldsymbol{\gamma}_{01,k},
\gamma_{10,k,0},\boldsymbol{\gamma}_{10,k}
&\sim
N(0,1).
\end{align}
For the loss transition, the intercept was assigned $\gamma_{10,k,0}\sim N(-2,1)$, while the covariate coefficients $\boldsymbol{\gamma}_{10,k}$ were assigned $N(0,1)$.

\subsection{Joint Bayesian Estimation}

The joint Bayesian procedure estimates the measurement model, the unknown $Q$-matrix, the latent mastery trajectories, item parameters, and structural regression parameters simultaneously. Let $Y=\{Y_1,\ldots,Y_T\}$ denote the observed item responses across all time points, and let $Z$ denote the collection of covariates used in the initial mastery and transition models. Let $\boldsymbol{\tau}=\{\boldsymbol{\tau}_1,\ldots,\boldsymbol{\tau}_T\}$ denote the item-level text-derived signals, where $\boldsymbol{\tau}_t=(\tau_{1t},\ldots,\tau_{J_t t})^{\top}$ at time $t$. The joint posterior distribution is proportional to
\begin{align}
& p(
Q_{1:T},
\boldsymbol{\alpha}_{1:T},
\mathbf g_{1:T},
\mathbf s_{1:T},
\boldsymbol{\beta},
\boldsymbol{\gamma},
\theta,
\lambda
\mid
Y,Z,\boldsymbol{\tau}
)
\nonumber \\
&\quad \propto
\prod_{t=1}^{T}
p(Y_t\mid \boldsymbol{\alpha}_t,Q_t,\mathbf g_t,\mathbf s_t)
p(Q_t\mid \theta,\lambda,\boldsymbol{\tau}_t)
\nonumber\\
&\qquad \times
\prod_{i=1}^{N}
p(\boldsymbol{\alpha}_{i1}\mid Z_{i0},\boldsymbol{\beta})
\prod_{t=2}^{T}
p(\boldsymbol{\alpha}_{it}\mid \boldsymbol{\alpha}_{i,t-1},Z_{i,t-1},\boldsymbol{\gamma})
\nonumber\\
&\qquad \times
p(\mathbf g_{1:T})
p(\mathbf s_{1:T})
p(\boldsymbol{\beta})
p(\boldsymbol{\gamma})
p(\theta)
p(\lambda).
\end{align}

The first product contains the measurement likelihood and the text-informed prior for the unknown $Q$-matrix at each time point. The measurement likelihood is defined by the DINA response model, whereas $p(Q_t\mid \theta,\lambda,\boldsymbol{\tau}_t)$ denotes the constrained text-informed prior described above. The second line describes the structural model for initial mastery and subsequent transitions. Specifically, $p(\boldsymbol{\alpha}_{i1}\mid Z_{i0},\boldsymbol{\beta})$ gives the initial mastery distribution, and $p(\boldsymbol{\alpha}_{it}\mid \boldsymbol{\alpha}_{i,t-1},Z_{i,t-1},\boldsymbol{\gamma})$ gives the transition distribution from time $t-1$ to time $t$.

Posterior inference is conducted using Markov chain Monte Carlo sampling. At each iteration, the algorithm updates the latent mastery profiles, item parameters, structural regression parameters, text-informed prior parameters, and admissible $Q$-matrices. The admissibility restrictions on $Q_t$ are enforced throughout sampling, so posterior draws of the $Q$-matrix always satisfy the structural conditions described above. Posterior summaries, including posterior means, credible intervals, and classification probabilities, are used to evaluate item-attribute relationships, student mastery trajectories, and covariate effects on learning transitions.

\subsection{Bias-Corrected Stepwise Estimation}

The stepwise strategy separates measurement estimation from structural estimation. The first step estimates the measurement model separately at each time point. The second step converts posterior mastery probabilities into hard classifications and estimates classification error probabilities. The third step fits the dynamic structural model using those hard classifications while correcting for classification error.

\subsubsection{Step 1: Measurement Model Estimation}

At each time point, the following measurement posterior is estimated independently:
\begin{align}
p(\boldsymbol{\alpha}_t,Q_t,\mathbf g_t,\mathbf s_t,\theta,\lambda\mid Y_t,\mathbf T_t)
&\propto
p(Y_t\mid \boldsymbol{\alpha}_t,Q_t,\mathbf g_t,\mathbf s_t)
p(Q_t\mid \theta,\lambda,\mathbf T_t) \nonumber\\
&\quad \times
p(\boldsymbol{\alpha}_t)
p(\mathbf g_t)p(\mathbf s_t)p(\theta)p(\lambda).
\end{align}

The first-step measurement model is estimated independently of the longitudinal structural model. This step can be understood as fitting a cross-sectional cognitive diagnostic model. A cognitive diagnostic model is a restricted latent class model in which each latent class corresponds to an attribute mastery profile. For example, with $K$ binary attributes, student $i$’s mastery profile $\boldsymbol{\alpha}_{it}$ belongs to one of $2^K$ possible latent classes. In a standard latent class formulation, the measurement model requires a population distribution over these latent classes, such as
\begin{align*}
P(\boldsymbol{\alpha}_{it}=c)=\pi_{ct}, \quad c=1,\ldots,2^K.
\end{align*} 

This latent class distribution describes how likely each mastery profile is in the student population at time $t$. In the present implementation, we use an attribute-wise version of this idea. Instead of assigning a probability to each full mastery profile, we assign a marginal mastery probability to each attribute:
\begin{align}
\alpha_{ikt}\mid \rho_{kt}
&\sim
\mathrm{Bernoulli}(\rho_{kt}), \\
\rho_{kt}
&\sim
\mathrm{Beta}(a_{\rho},b_{\rho}).
\end{align}
In the empirical stepwise analysis, we set $a_{\rho}=1$ and $b_{\rho}=1$. The parameter $\rho_{kt}\in (0,1)$ represents the marginal probability that a student has mastered attribute $k$ at time $t$ in the first-step measurement model. This parameter is not interpreted as a learning, acquisition, or transition parameter. The learning process is modeled later through the corrected structural model.

\subsubsection{Step 2: Hard Classification and Classification Error Probabilities}

Let
\begin{align}
\widehat{p}_{ikt}
&=
P(\alpha_{ikt}=1\mid Y_t)
\end{align}
denote the posterior probability of mastery from the first-step measurement model. The hard assigned class is
\begin{align}
W_{ikt}
&=
I(\widehat{p}_{ikt}\geq 0.5).
\end{align}
The assigned class $W_{ikt}$ is treated as an observed but error-prone indicator of the unobserved true mastery state, denoted $L_{ikt}$.

For each attribute $k$ and time $t$, the classification error probability matrix is
\begin{align}
M_{kt}(w,l)
&=
P(W_{ikt}=w\mid L_{ikt}=l),
\qquad
w,l\in\{0,1\}.
\end{align}
Its entries are estimated from the posterior probabilities obtained in Step 1:
\begin{align}
\widehat{M}_{kt}(w,l)
&=
\frac{
\sum_{i=1}^{N}
P(L_{ikt}=l\mid Y_t)I(W_{ikt}=w)
}{
\sum_{i=1}^{N}
P(L_{ikt}=l\mid Y_t)
}.
\end{align}
Thus, if a student is assigned $W_{ikt}=0$, the structural model evaluates how compatible that assignment is with both possible true states through $\widehat{M}_{kt}(0,0)$ and $\widehat{M}_{kt}(0,1)$. If a student is assigned $W_{ikt}=1$, the corresponding correction probabilities are $\widehat{M}_{kt}(1,0)$ and $\widehat{M}_{kt}(1,1)$.

\subsubsection{Step 3: Corrected Structural Model}

In the third step, the structural model is fitted to the assigned classifications $W$, but the likelihood marginalizes over the possible true latent states $L$. For two occasions and one attribute, the corrected likelihood contribution is
\begin{align}
P(W_{ik1}=w_1,W_{ik2}=w_2\mid Z_i)
&=
\sum_{l_1=0}^{1}
\sum_{l_2=0}^{1}
P(L_{ik1}=l_1\mid Z_{i0},\boldsymbol{\beta}_k) \nonumber\\
&\quad \times
P(L_{ik2}=l_2\mid L_{ik1}=l_1,Z_{i1},\boldsymbol{\gamma}_k)
\widehat{M}_{k1}(w_1,l_1)
\widehat{M}_{k2}(w_2,l_2).
\end{align}
For example, when the observed assigned trajectory is $(W_{ik1},W_{ik2})=(1,1)$, the likelihood is the sum of four possible true trajectories:
\begin{align}
P(W_{ik1}=1,W_{ik2}=1\mid Z_i)
&=
P(L_{ik1}=0\mid Z_{i0})
P(L_{ik2}=0\mid L_{ik1}=0,Z_{i1})
\widehat{M}_{k1}(1,0)\widehat{M}_{k2}(1,0) \nonumber\\
&\quad+
P(L_{ik1}=0\mid Z_{i0})
P(L_{ik2}=1\mid L_{ik1}=0,Z_{i1})
\widehat{M}_{k1}(1,0)\widehat{M}_{k2}(1,1) \nonumber\\
&\quad+
P(L_{ik1}=1\mid Z_{i0})
P(L_{ik2}=0\mid L_{ik1}=1,Z_{i1})
\widehat{M}_{k1}(1,1)\widehat{M}_{k2}(1,0) \nonumber\\
&\quad+
P(L_{ik1}=1\mid Z_{i0})
P(L_{ik2}=1\mid L_{ik1}=1,Z_{i1})
\widehat{M}_{k1}(1,1)\widehat{M}_{k2}(1,1).
\end{align}
For multiple time points, this correction generalizes to
\begin{align}
P(W_{ik1},\ldots,W_{ikT}\mid Z_i)
&=
\sum_{l_1=0}^{1}\cdots\sum_{l_T=0}^{1}
P(L_{ik1}=l_1\mid Z_{i0},\boldsymbol{\beta}_k) \nonumber\\
&\quad \times
\prod_{t=2}^{T}
P(L_{ikt}=l_t\mid L_{ik,t-1}=l_{t-1},Z_{i,t-1},\boldsymbol{\gamma}_k)
\prod_{t=1}^{T}
\widehat{M}_{kt}(W_{ikt},l_t).
\end{align}
The stepwise posterior for the structural parameters is therefore
\begin{align}
p(\boldsymbol{\beta},\boldsymbol{\gamma}\mid W,Z,\widehat{M})
&\propto
\prod_{i=1}^{N}
\prod_{k=1}^{K}
P(W_{ik1},\ldots,W_{ikT}\mid Z_i,\widehat{M}_{k1},\ldots,\widehat{M}_{kT})
p(\boldsymbol{\beta})p(\boldsymbol{\gamma}).
\end{align}

\subsection{Comparison of the Two Strategies}

The two approaches differ in how they condition on the measurement model. The joint approach targets
\begin{align}
p(
\boldsymbol{\alpha}_{1:T},
Q_{1:T},
\mathbf g_{1:T},
\mathbf s_{1:T},
\boldsymbol{\beta},
\boldsymbol{\gamma}
\mid
Y,Z,T
),
\end{align}
so learning parameters are estimated while accounting for posterior uncertainty in the latent states, item parameters, and $Q$-matrix. The stepwise approach targets
\begin{align}
p(\boldsymbol{\beta},\boldsymbol{\gamma}\mid W,Z,\widehat{M}),
\end{align}
where $W$ and $\widehat{M}$ are functions of the first-step measurement posterior. Thus, the stepwise approach preserves the separation between measurement and structural modeling, while the joint approach estimates all components simultaneously.

This distinction is important for interpretation. In the joint strategy, an estimated acquisition effect reflects evidence from the full response process, the text-informed $Q$-matrix prior, item-level uncertainty, and the dynamic structural model at the same time. In the stepwise strategy, an estimated acquisition effect reflects the association between covariates and the corrected assigned mastery trajectories after the measurement model has already been established. When the measurement model is strong and classification error is small, the two approaches should provide similar conclusions. When the $Q$-matrix is uncertain, tests are short, or classifications are unstable, the two strategies may lead to different interpretations of learning.

\subsubsection{Implementation}

The stepwise procedure was implemented in three stages. First, the DINA measurement model was estimated separately at each time point using \texttt{nimble}. The $Q$-matrix was treated as unknown and estimated jointly with the item parameters and latent attribute indicators in the first-step measurement model. Second, posterior mastery probabilities from the first-step model were used to obtain hard classifications and to estimate the classification error probability matrices. Third, the structural transition model was fitted using a Bayesian implementation of the corrected marginal likelihood.

This implementation preserves the bias-corrected logic of the original stepwise approach \citep{liang2023latent}. The measurement and structural components remain separated, and classification error is corrected through the estimated classification error probability matrices. Compared with the earlier work, the present implementation extends the stepwise procedure to the case where the $Q$-matrix is unknown. The measurement model is estimated in \texttt{nimble} with the $Q$-matrix treated as unknown, and the structural parameters are estimated using Bayesian posterior sampling rather than maximum likelihood optimization.

The joint procedure was implemented by estimating the measurement model, unknown $Q$-matrix, latent mastery trajectories, item parameters, and structural transition parameters simultaneously within a single Bayesian model. Unlike the stepwise procedure, the joint procedure does not fix first-step mastery classifications or rely on a separate classification error correction. Instead, uncertainty in the $Q$-matrix, item parameters, and latent mastery states is estimated directly into posterior inference for the structural transition parameters. The code for implementing the proposed framework is available on the Open Science Framework (OSF) \footnote{Code is available at: \url{https://osf.io/tjqbp/overview?view_only=b3672f24ba8f43ebb99f86f2d0ec8ef4}.}, and an accompanying R package is under development.

\section{Empirical Study}

The empirical study used log files provided by Amplify from the Boost Reading platform (\url{https://amplify.com/programs/boost-reading/}). We focused on two reading-related games: \textit{Punchline} and \textit{Field Observer}. These two games were selected because they target different reading skills and provide repeated item responses that can be linked to specific attributes. \textit{Punchline} mainly targets vocabulary knowledge, whereas \textit{Field Observer} mainly targets comprehension skills related to key ideas and details. These games also focus on critical comprehension-related skills that predict reading comprehension in this age group, when word recognition skills are becoming fluent \citep{garcia2014decoding, oakhill2012precursors}.

\subsection{Game Description and Item Selection}

\textit{Punchline} is a vocabulary game based on jokes and word meanings. The game contains 11 levels. Each level includes several questions related to words with multiple meanings. In the original game design, students complete two response steps. First, they choose between two sentence options and identify the option whose meaning is most consistent with the question. Second, after selecting the sentence, they identify the target word that carries two different meanings. The game also includes an instructional component in which students are shown explanations of the two meanings of the target word. For the present analysis, only the first response step was used. That is, each analyzed response corresponded to choosing the correct option from two alternatives. The second response step, in which students selected the word with two meanings from the chosen sentence, was not included in the scored item responses because this learning step was not consistently represented in the scoring structure used for the analysis. Each level contained six analyzed questions, with every two questions associated with one relevant topic. Thus, each level covered three target words, and the analytic responses reflected whether students selected the sentence option that matched the intended meaning in context.

\textit{Field Observer} is a comprehension game focused on key ideas and details. The game contains 12 levels. In this game, students search for hidden animals and answer comprehension questions. The order in which animals are found is not fixed, so the order of questions may vary across students. For each question, students are presented with a question and a set of evidence statements. They select an answer option and then map the answer back to the corresponding evidence. In the original game design, a correct response requires both selecting the correct answer and identifying the corresponding supporting evidence. In the present analysis, the item was represented using the question together with the true evidence as the item stem, and the response reflected whether the student selected the correct answer. Distractor evidence was not used in constructing the item text for the text-informed prior. Therefore, for \textit{Field Observer}, each analyzed item was defined by the combination of the question and the true supporting evidence.

For both games, we analyzed item responses from the same cohort of students at two time points: U.S. Grade 2 and Grade 3. To ensure that the same students had sufficient participation at both time points, we conducted exploratory data analysis to balance sample size and item length (\ref{AppendixA}). Based on this trade-off, the final sample contained 1,978 students. At each time point, the model used 12 selected items: six \textit{Punchline} items and six \textit{Field Observer} items. Across the two time points, this corresponded to 24 items. These items were used to estimate the latent mastery profiles and learning transitions in the proposed framework.

\subsection{Descriptive Statistics}

The descriptive statistics for the categorical and continuous covariates are summarized in Tables~\ref{tab:summary_categorical} and~\ref{tab:summary_continuous}. As shown in Table~\ref{tab:summary_categorical}, a large proportion of students were classified as ``Above Benchmark'' or ``At Benchmark'' based on their initial DIBELS benchmark level. This pattern is partly consistent with the selected item set, which consisted of Grade 2 and Grade 3 content. Students classified as ``Below Benchmark'' or ``Well Below Benchmark'' may have been less likely to encounter these items if the adaptive platform placed them in earlier or below-grade content.

The categorical covariates contained substantial not-specified responses for several demographic variables, particularly race, SEN, and ELL. The gender distribution included a higher proportion of male students than female students. The sample also included students from diverse racial and ethnic backgrounds, including White, Asian, Black or African-American, Multiracial, American Indian or Alaska Native, Native Hawaiian or Other Pacific Islander, Other, and not specified categories.

Table~\ref{tab:summary_continuous} presents summary statistics for the log-derived behavioral variables. On average, students used more attempts in \textit{Punchline} than in \textit{Field Observer}. Performance on \textit{Punchline} showed greater variability, whereas students answered nearly all selected \textit{Field Observer} items correctly on average. Average response time was shorter for \textit{Punchline} than for \textit{Field Observer}, which is consistent with the comprehension-based structure of \textit{Field Observer}, where students needed to process both a question and supporting evidence.

\begin{table}[ht]
\centering
\caption{Summary of Categorical Variables.\\
\textit{Note: SEN = special educational needs; ELL = English language learner. Missing indicates students without matched demographic records}}
\label{tab:summary_categorical}
\begin{tabular}{ll}
\hline
\textbf{Variable} & \textbf{Summary} \\
\hline
\multicolumn{2}{l}{\textit{Demographic Variables}} \\
\hline
Race & White: 624 (31.55\%); Asian: 198 (10.01\%); \\
     & Black or African-American: 182 (9.20\%); Multiracial: 143 (7.23\%); \\
     & American Indian or Alaska Native: 7 (0.35\%); \\
     & Native Hawaiian or Other Pacific Islander: 3 (0.15\%); \\
     & Other: 44 (2.22\%); Not specified: 770 (38.93\%); \\
     & Missing: 7 (0.35\%) \\
SEN & Non-SEN: 915 (46.26\%); SEN: 69 (3.49\%); \\
    & Not specified: 987 (49.90\%); Missing: 7 (0.35\%) \\

ELL & Not ELL: 932 (47.12\%); ELL: 155 (7.84\%); \\
    & Not specified: 884 (44.69\%); Missing: 7 (0.35\%) \\

Gender & Female: 710 (35.89\%); Male: 1021 (51.62\%); \\
       & Not specified: 240 (12.13\%); Missing: 7 (0.35\%) \\
\hline
\multicolumn{2}{l}{\textit{Initial Literacy Ability from DIBELS Scores}} \\
\hline
Initial Literacy Ability & Above Benchmark: 1374 (69.46\%); At Benchmark: 459 (23.21\%); \\
                         & Below Benchmark: 83 (4.20\%); Well Below Benchmark: 62 (3.13\%) \\
\hline
\end{tabular}
\end{table}

\begin{table}[ht]
\centering
\caption{Summary of Continuous Variables.\\
\textit{Note: Response time is summarized using the transformed response-time variable used in the analysis}}
\label{tab:summary_continuous}
\begin{tabular}{lcccc}
\hline
\textbf{Covariate} & \textbf{Mean} & \textbf{SD} & \textbf{Median} & \textbf{Min, Max} \\
\hline
Average Attempts (\textit{Punchline})      & 1.84 & 1.14 & 2.00 & 1.00, 14.00 \\
Average Attempts (\textit{Field Observer}) & 1.39 & 1.01 & 1.00 & 1.00, 16.00 \\
Correct Answers (\textit{Punchline})       & 4.26 & 1.49 & 4.00 & 0.00, 6.00 \\
Correct Answers (\textit{Field Observer})  & 5.75 & 0.52 & 6.00 & 2.00, 6.00 \\
Response Time (\textit{Punchline})         & 2.00 & 1.18 & 1.79 & 0.76, 23.33 \\
Response Time (\textit{Field Observer})    & 5.05 & 2.25 & 4.64 & 1.27, 45.44 \\
\hline
\end{tabular}
\end{table}

\subsection{Empirical Comparison of Joint and Stepwise Results}

The empirical comparison focuses on the text-informed $Q$-matrix prior because the main inferential question is how the joint and stepwise strategies support different interpretations of learning when the measurement structure is uncertain. The goal of the empirical analysis is not to isolate the incremental contribution of the text-derived prior relative to a non-text prior. Rather, the text-informed prior provides a common measurement framework under which the joint and stepwise strategies can be compared using the same available item-level information. Both empirical analyses used three Markov chains with 10,000 burn-in iterations and 10,000 monitored iterations. In the joint text-prior model, the posterior mean of $\lambda$ was 0.075, with posterior standard deviation 0.264 and a 95\% credible interval of $(-0.440,0.561)$. In the stepwise text-prior measurement models, the estimated $\lambda$ was -0.164 at time 1 and 0.170 at time 2. 

Table \ref{tab:emp_q_item_joint_stepwise} places the text-prior joint and stepwise $Q$-matrix and item-parameter estimates in the same table. The two approaches agree on several item-attribute patterns, especially for high-performing items at Time 2, but they also produce different $Q$-matrix rows for a number of items. These differences are important because they show that the interpretation of the latent attributes can depend on whether measurement uncertainty is incorporated jointly through the dynamic model or resolved first in separate measurement steps.

\input{empirical_tables/table_emp_q_item_joint_stepwise.tex}

Table \ref{tab:emp_profile_transition_joint_stepwise} reports the profile transition summaries for the text-prior joint and stepwise approaches. Both approaches indicate movement toward full mastery of both attributes at time 2. The joint model assigns 1,730 students to profile $(1,1)$ at time 2, whereas the stepwise classification summary assigns 1,659 students to this profile. The stepwise summary also places more students in the time 2 profile $(0,1)$ than the joint model. Because the stepwise transition table is based on hard classifications from the separate first-step measurement models, these differences should be interpreted as differences in the empirical classification pathway rather than as direct discrepancies in a single common posterior distribution. Substantively, both analyses suggest that most students showed evidence of broad mastery by the second time point. The difference between the two approaches is therefore not a disagreement about whether students generally moved toward mastery, but about how confidently students should be assigned to full mastery rather than partial mastery at Grade 3. Under the joint model, the interpretation is broad consolidation of both attributes. Under the stepwise model, a larger subgroup remains classified as having mastered only one of the two attributes. This distinction matters because it affects whether the empirical learning process is interpreted primarily as general movement toward full mastery or as a more heterogeneous pattern in which some students remain partially mastered at the second occasion.

\input{empirical_tables/table_emp_profile_transition_joint_stepwise.tex}

For an applied reading programme, this difference is more than statistical. The two analyses would support different summaries of the same cohort. Under the joint model, most children appear to have consolidated both vocabulary and comprehension by Grade 3, and a progress report based on it would describe near-uniform mastery. Under the stepwise model, a larger group is credited with only one of the two skills, and a report based on it would flag a sizeable minority as still developing one component. The underlying data are identical; what differs is whether uncertainty in the measurement model is carried into the classification or resolved before it. A programme deciding who to target for additional instruction would identify a different set of children under each analysis.

Table \ref{tab:emp_significant_covariates_joint_stepwise} summarizes statistically credible covariate effects for initial mastery and acquisition transitions under both approaches. For initial mastery, both methods identify the number of correctly answered \textit{Punchline} questions as a strong positive predictor and the number of attempts in \textit{Punchline} as a negative predictor. For acquisition transitions, both methods identify the number of correctly answered \textit{Field Observer} questions as a positive predictor and ``Well Below Benchmark'' status as a negative predictor. The stepwise model additionally identifies negative effects for the number of correctly answered \textit{Punchline} questions and the number of attempts in \textit{Field Observer} on acquisition of comprehension skill, as well as a positive effect of ``Above Benchmark'' status for comprehension skill. These results suggest that the most stable empirical signals are associated with behavioural indicators from the learning environment and initial literacy benchmark status, rather than with demographic variables. For initial mastery, the positive effect of correctly answered \textit{Punchline} questions and the negative effect of \textit{Punchline} attempts are consistent with the interpretation that students who answered more vocabulary-related items correctly and required fewer attempts were more likely to begin the observed period with stronger mastery. For acquisition transitions, the positive effect of correctly answered \textit{Field Observer} questions and the negative effect of ``Well Below Benchmark'' status indicate that comprehension-related task performance and baseline literacy status are the clearest markers of subsequent mastery acquisition.

The additional effects identified only by the stepwise model should be interpreted cautiously. Because the stepwise approach first converts posterior mastery probabilities into hard classifications and then corrects for classification error, some covariate effects may reflect the specific classification pathway induced by the first-stage measurement models. These stepwise-only effects may indicate that the stepwise model distinguishes more sharply between students who were already high performing and students whose response patterns provide evidence of transition into mastery. In contrast, the joint model carries uncertainty in the $Q$-matrix, item parameters, and latent states directly into the transition model, leading to a smaller set of acquisition effects that are supported across the full dynamic measurement process.

\input{empirical_tables/table_emp_significant_covariates_joint_stepwise.tex}

Detailed posterior odds ratios and 95\% credible intervals are reported in Tables \ref{tab:emp_betaZ_OR_part1_joint_stepwise}-\ref{tab:emp_gamma01_OR_part2_joint_stepwise}. Tables \ref{tab:emp_betaZ_OR_part1_joint_stepwise} and \ref{tab:emp_betaZ_OR_part2_joint_stepwise} report covariate effects on initial mastery, with joint rows followed by stepwise rows. Tables \ref{tab:emp_gamma01_OR_part1_joint_stepwise} and \ref{tab:emp_gamma01_OR_part2_joint_stepwise} report acquisition transition effects in the same format. Across these detailed tables, the strongest empirical signals are concentrated in log-derived behavioral variables and benchmark categories, while demographic effects are generally more uncertain.

\input{empirical_tables/table_emp_betaZ_OR_part1_joint_stepwise.tex}
\input{empirical_tables/table_emp_betaZ_OR_part2_joint_stepwise.tex}
\input{empirical_tables/table_emp_gamma01_OR_part1_joint_stepwise.tex}
\input{empirical_tables/table_emp_gamma01_OR_part2_joint_stepwise.tex}

\section{Simulations}
\subsection{Simulation Design}
The simulation study was designed to reflect the scale of the empirical application while also evaluating the model under varying levels of measurement information. For each simulation condition, we conducted 100 independent replications. We considered a dynamic CDM with two attributes measured across two time points. Sample size varied across conditions as $N \in \{1000, 2000, 4000\}$, and the number of items administered at each time point $t$ varied as $J_t \in \{6, 12, 24\}$. The middle condition, $N=2000$ and $J_t=12$ at each time point, was chosen to approximate the empirical setting. The true $Q$-matrices used under each simulation condition are provided in Table \ref{tab:qmatrix_j24_t1_t2} in the Appendix. The prior for $\theta$ was specified as $\mathrm{Beta}(6,4)$, with a prior mean of $0.6$. The prior for $\lambda$ was specified as $\mathcal{N}(0,0.5^2)$.

\subsection{Simulation Results}

The simulation results are summarized in Tables \ref{tab:q_recovery_joint_stepwise}-\ref{tab:regression_recovery_joint_stepwise}. Each fit used three independent Markov chains, with 3,000 burn-in iterations followed by 3,000 monitored iterations. Standard errors in the tables were computed using 1,000 nonparametric bootstrap resamples over successful replications.

The average running time per text-prior fit was approximately 33.1 to 216.0 minutes for the joint approach and 21.5 to 119.4 minutes for the stepwise approach across different simulation settings. These runs were conducted on a MacBook Pro (13-inch, M1, 2020) equipped with an Apple M1 chip (8-core: 4 performance and 4 efficiency cores) and 16 GB of unified memory.

Table \ref{tab:q_recovery_joint_stepwise} reports recovery of the time-specific $Q$-matrices. The joint model recovered the $Q$-matrix nearly perfectly across the successful conditions, with entry-level ACC values equal or close to one and very small PIP errors. The stepwise approach also recovered $Q_1$ well, with ACC ranging from 0.917 to 1.000, but recovery of $Q_2$ was less stable. Stepwise $Q_2$ ACC ranged from 0.593 to 0.843 across reported conditions, indicating that uncertainty in the second time point was more difficult to recover in the separate first-stage measurement models.

\input{simu_tables/table_Q_recovery_joint_stepwise.tex}

Table \ref{tab:alpha_recovery_joint_stepwise} presents recovery of latent attribute profiles. The joint model achieved high profile agreement rates at both time points, with PAR values ranging from 0.926 to 0.989 for $\alpha_1$ and from 0.923 to 0.954 for $\alpha_2$ among reported conditions. The stepwise approach gave comparable recovery for $\alpha_1$ in some conditions, but recovery of $\alpha_2$ was consistently weaker, with PAR ranging from 0.668 to 0.779. These results suggest that classification error from the separate measurement steps can affect the transition stage, whereas the joint model stabilizes the two time points by estimating measurement and transition components simultaneously.

\input{simu_tables/table_alpha_recovery_joint_stepwise.tex}

Table \ref{tab:item_recovery_joint_stepwise} summarizes guessing and slipping parameter recovery. Both approaches recovered item parameters accurately when the stepwise model converged. Item-parameter RMSE and MAE were small across conditions, and they generally decreased as sample size increased. Differences between the joint and stepwise approaches were much smaller for item parameters than for transition-related quantities. This suggests that the largest practical distinction between the two frameworks is not in estimating the DINA measurement parameters once the measurement model is identifiable, but rather in how uncertainty about the $Q$-matrix and latent classifications is incorporated into later components of the dynamic model.

\input{simu_tables/table_item_recovery_joint_stepwise.tex}

Table \ref{tab:regression_recovery_joint_stepwise} reports estimation accuracy for the initial mastery parameters $\beta_0$ and $\beta_Z$ and the acquisition transition parameters $\gamma_{01}$. The joint model consistently produced smaller errors for $\gamma_{01}$ than the stepwise approach in the reported conditions. For example, under $N=2000,J_t=12$, the joint $\gamma_{01}$ RMSE was 0.080, compared with 0.326 for the stepwise model. Under $N=2000,J_t=24$, the corresponding RMSE values were 0.099 and 0.428. Differences for $\beta_0$ and $\beta_Z$ were smaller, but the transition parameters showed a clear advantage for joint estimation. This is consistent with the expectation that transition effects are particularly sensitive to classification uncertainty in latent mastery profiles.

\input{simu_tables/table_regression_recovery_joint_stepwise.tex}

Overall, the simulation results indicate that joint estimation is more robust when the $Q$-matrix is unknown and measurement information is limited. The stepwise procedure can give reasonable item-parameter estimates when the first-step measurement model is identified by the data, but it performs worse than the joint strategy for the second time point and for transition parameters. This pattern is consistent with the role of the joint model in carrying uncertainty about $Q$-matrix recovery and latent mastery classifications directly into the dynamic transition component.

\section{Discussion}

This study compared joint and stepwise estimation strategies for longitudinal CDMs when the $Q$-matrix is unknown and informed by item text information. Across the simulation conditions, the joint approach generally produced more stable recovery of latent attribute profiles, transition parameters, and covariate effects, particularly when the number of items was limited or when classification uncertainty increased over time. By contrast, the stepwise approach often produced reasonable recovery for measurement parameters at the first time point, but the performance is weaker than that of the joint approach for the second time point and for the longitudinal transition components.

The distinction between the two strategies is therefore not merely computational, but also interpretive. The joint approach treats the measurement model, latent transitions, covariate effects, and observed responses as components of a single probabilistic system that are estimated simultaneously. In contrast, the stepwise approach separates measurement from longitudinal modelling by first estimating latent profiles and then using those identified profiles in later stages of analysis. When measurement information is sufficiently strong, both approaches can lead to similar conclusions. However, the two strategies may diverge because uncertainty from earlier measurement stages affects subsequent transition estimation differently.

The results further suggest that the largest practical distinction between the two frameworks is not primarily in estimating DINA item parameters themselves, but rather in how uncertainty surrounding the $Q$-matrix and latent mastery profiles is incorporated into later components of the longitudinal model. Under the stepwise framework, classification decisions from earlier stages may introduce instability into transition and regression parameters estimation when those classifications are uncertain. The joint framework instead allows measurement and transition information to inform one another simultaneously, leading to more coherent estimation across time points.

For applied reading assessment, both strategies agree that children in this cohort generally progress toward mastering vocabulary and comprehension between Grade 2 and Grade 3. They differ in how many children are reported as having fully consolidated both skills, and that difference is largest exactly where digital reading data are weakest: short item sets, item pools that change between grades, and an unknown attribute structure. In such settings, a programme that classifies children for targeted support should be aware that a stepwise approach can place more children in partial-mastery categories than a joint analysis of the same data, not because more children are struggling, but because uncertainty resolved early is read as a definite gap. Where the measurement model is strong the choice is inconsequential; where it is not, the joint analysis gives the more defensible basis for reporting and for deciding who needs further support.

These findings also highlight the importance of carefully interpreting the $Q$-matrix in longitudinal settings. In many educational applications, the assumption that the same attributes are measured identically over time may not fully hold because items, task formats, or instructional emphases can change across grades or school years. In the present study, the item pools differed across time points, making the recovery of attribute structures particularly important for maintaining interpretability of longitudinal transitions. This issue becomes increasingly relevant in digital learning environments, where item content evolves dynamically and item pools may be continuously updated.

In the empirical analysis, item text contributed little to the estimated structure. It entered as a prior on the $Q$-matrix, and the posterior for $\lambda$ was close to zero, so the response data effectively determined the estimated structure. We do not read this as
evidence that item text is uninformative. The responses in these data were strong enough to identify the $Q$-matrix on their own, leaving the prior with little to do. The value of the text component here is that it lets the joint and stepwise strategies draw on the same item information, so the two can be compared on equal
terms. \cite{ma2026nlp} show that the text-informed prior contributes more when the item response signal is weaker. 

Several limitations should also be noted. First, the proposed Bayesian estimation procedure relies on Markov chain Monte Carlo sampling and can be computationally intensive, particularly under the joint framework and larger item pools. Second, the empirical application was based on a selected subset of frequently administered items rather than the full item pool. Consequently, the findings may depend partly on the characteristics of the selected items and the resulting sample restrictions. Third, the current study focused on the DINA model with binary attributes. Future research could extend the framework to higher-order or more complex CDMs that allow hierarchical or continuous latent structures. Finally, the log-derived behavioural covariates used in the empirical study may themselves contain measurement error, which was not explicitly modelled in the current analysis. Incorporating uncertainty in process-based covariates may further improve inference in longitudinal diagnostic modelling.

Overall, the results suggest that joint estimation provides a more robust framework for longitudinal cognitive diagnosis when the $Q$-matrix is uncertain and measurement information is limited. Although stepwise procedures may remain attractive because of their flexibility and lower computational burden, the joint framework offers a more coherent representation of uncertainty across measurement and learning processes over time.

\newpage

\bibliography{bibliography}

\newpage

\section{Appendix A. Data Preprocessing}\label{AppendixA}

The empirical data were obtained from four game-grade combinations: Grade 2 \textit{Punchline}, Grade 3 \textit{Punchline}, Grade 2 \textit{Field Observer}, and Grade 3 \textit{Field Observer}. The total number of students in each dataset was as follows: 27,649 for Grade 2 \textit{Punchline}, 4,593 for Grade 3 \textit{Punchline}, 13,044 for Grade 2 \textit{Field Observer}, and 18,220 for Grade 3 \textit{Field Observer}. To construct a longitudinal sample, we identified students who appeared in both Grade 2 and Grade 3 datasets. 

We conducted item selection separately for each game and grade based on student participation. For each game and grade, we computed (a) the number of students who attempted each item and (b) the proportion of participating students. Items were then ranked according to their frequency of appearance to identify those with the highest coverage. For each game-grade combination, we selected the six most frequently attempted items after balancing the trade-off between including more items and retaining a larger sample size. This produced 12 selected items at each grade, with six items from \textit{Punchline} and six items from \textit{Field Observer}.

The selected items were:
\begin{itemize}
    \item Grade 2 \textit{Punchline}: 1,2,3,7,8,9
    \item Grade 3 \textit{Punchline}: 1,2,3,31,32,33
    \item Grade 2 \textit{Field Observer}: 1,3,5,7,9,11
    \item Grade 3 \textit{Field Observer}: 31,33,35,37,39,41
\end{itemize}

We then restricted the dataset to students who completed all selected items across both games and both time points (Grade 2 and Grade 3). This resulted in a final analytic sample of 1,978 students and ensured a balanced longitudinal structure for subsequent modelling.

\begin{figure}
    \centering
    \includegraphics[width=1\linewidth]{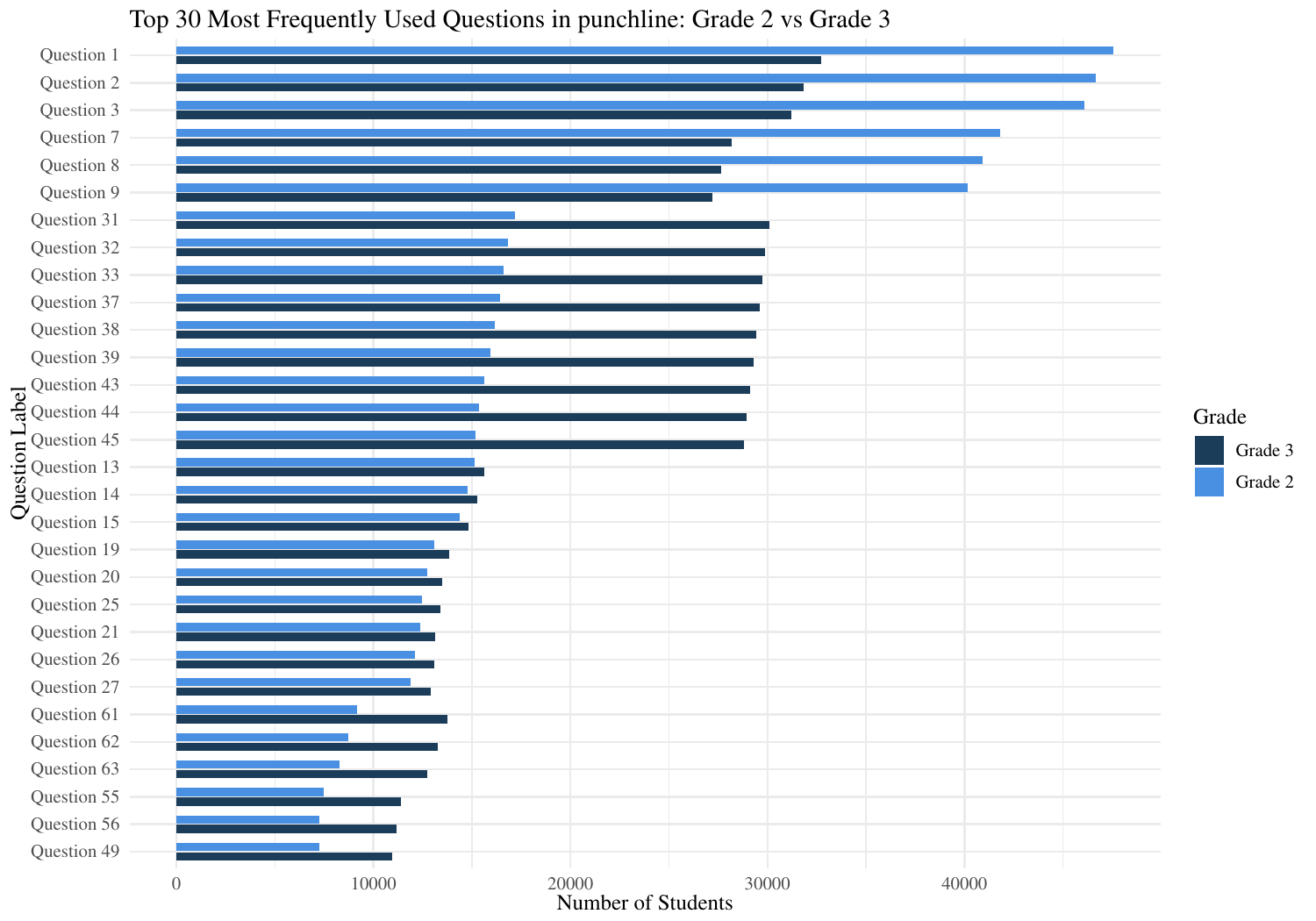}
     \caption{Top 30 most frequently attempted questions in punchline, separately for Grade 2 and Grade 3. Counts represent the number of students who attempted each question. Questions selected for the analysis are among the highest-frequency items}
    \label{fig:punchline_topgamesG12}
\end{figure}

\begin{figure}
    \centering
    \includegraphics[width=1\linewidth]{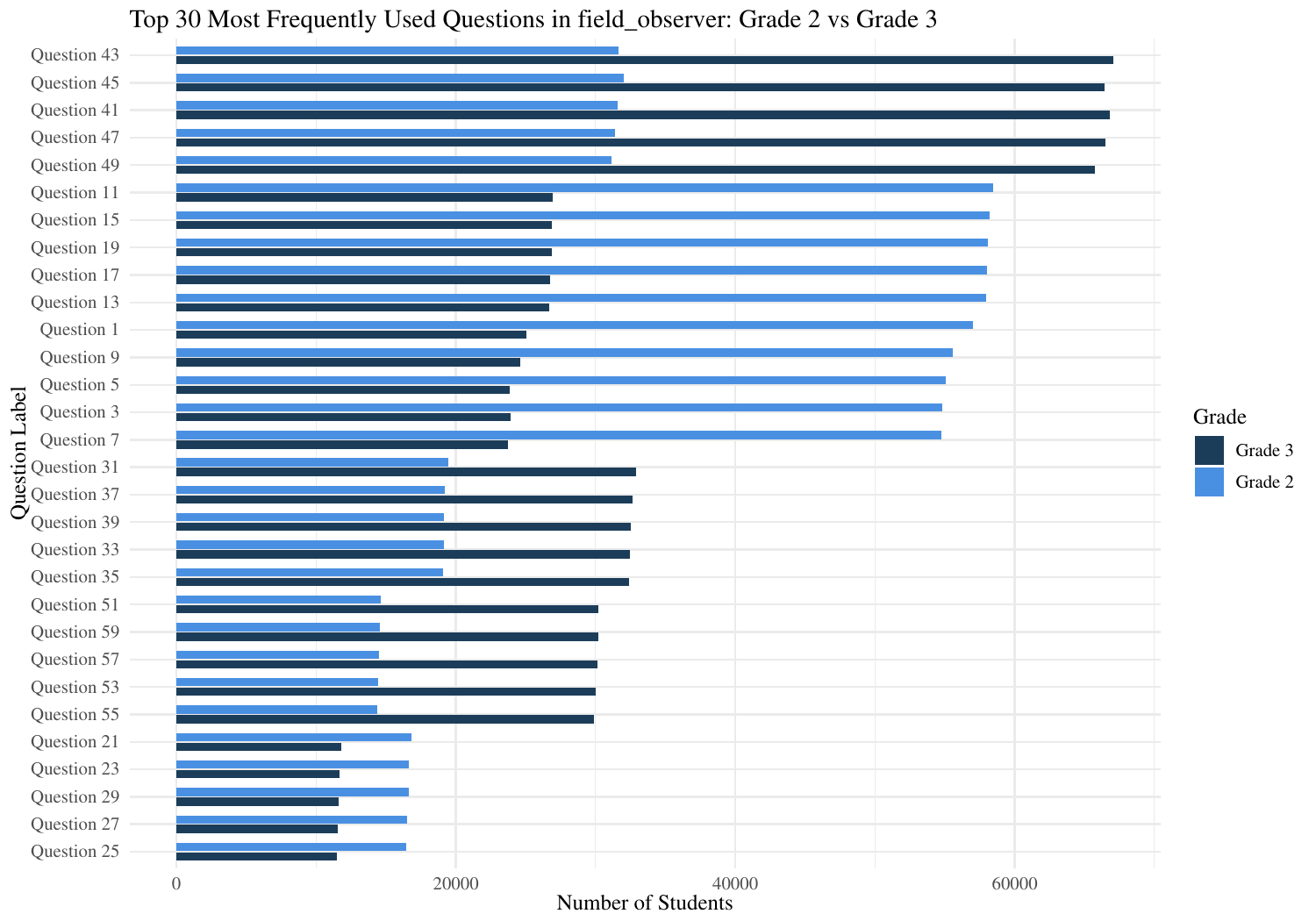}
    \caption{Top 30 most frequently attempted questions in field observer, separately for Grade 2 and Grade 3. Counts represent the number of students who attempted each question. Questions selected for the analysis are among the highest-frequency items}
    \label{fig:field_topgamesG12}
\end{figure}

To identify the most informative subset of items, we conducted an exploratory data analysis (EDA) on question usage frequency. For each question, the number and proportion of students who attempted the item were computed. The top 30 most frequently used questions in each game are shown in Figures \ref{fig:punchline_topgamesG12} and \ref{fig:field_topgamesG12}. Based on these distributions, a subset of high-frequency questions was selected to maximize sample size while maintaining sufficient coverage across levels.

The covariates were derived from students’ gameplay data across the two games. The number of attempts was computed as the average number of attempts per level for each student, and then averaged across the two games. The number of questions correct was defined as the total number of questions answered correctly across both games. Response time (RT) was calculated as the average response time per level for each student and subsequently averaged across the two games. These definitions and computation procedures are consistent with those used in the previous study by \cite{ma2026dynamic}.

\newpage

\section{Appendix B: Example of Questions}\label{AppendixB}

\begin{table}[H]
\centering
\caption{Example items from Punchline}
\label{tab:punchline_examples}
\setlength{\tabcolsep}{4pt}
\begin{tabular*}{\textwidth}
{@{\extracolsep{\fill}}p{0.08\textwidth}p{0.22\textwidth}p{0.66\textwidth}@{}}
\toprule
\textbf{Item} & \textbf{Component} & \textbf{Content} \\
\midrule

\multirow{5}{*}{1}
& Question & Why are pop stars so cool? \\
& Correct answer & They have lots of fans. \\
& Distractor answer & They make great music. \\
& Definition 1 & machines that make you cooler \\
& Definition 2 & people who are excited about someone or something \\
\midrule

\multirow{5}{*}{2}
& Question & Why did the little kitten run away from the tree? \\
& Correct answer & Because she saw the tree bark. \\
& Distractor answer & Because she thought the tree moved. \\
& Definition 1 & the rough outside of a tree trunk \\
& Definition 2 & the noise a dog makes \\
\midrule

\multirow{5}{*}{3}
& Question & How did Cinderella ruin the basketball game? \\
& Correct answer & She kept running away from the ball. \\
& Distractor answer & She left early. \\
& Definition 1 & a round object used to play games and sports \\
& Definition 2 & a big party where there is dancing \\
\bottomrule
\end{tabular*}

\vspace{0.5em}
\begin{minipage}{\textwidth}
\footnotesize
\textit{Note:} Each Punchline item requires students to select the correct answer and identify the relevant meaning of the ambiguous word.
\end{minipage}
\end{table}

\begin{table}[H]
\centering
\caption{Example items from Field Observer}
\label{tab:field_observer_examples}
\setlength{\tabcolsep}{4pt}
\begin{tabular*}{\textwidth}
{@{\extracolsep{\fill}}p{0.08\textwidth}p{0.20\textwidth}p{0.68\textwidth}@{}}
\toprule
\textbf{Item} & \textbf{Component} & \textbf{Content} \\
\midrule

\multirow{6}{*}{1}
& Question & What might happen if you asked this Curioso to carry a tray of water glasses? \\
& Relevant evidence & Coldocs always trip over furniture, fall down, and drop things. \\
& Distractor evidence & Coldocs only eat red foods, such as apples and red potatoes. \\
& Correct answer & He would likely break all the glasses. \\
& Distractor answer & He'd drink the water and wash the glasses. \\
& Distractor answer & He'd take it away without an accident. \\
\midrule

\multirow{6}{*}{2}
& Question & Should you leave anything in the Coldoc's cave? \\
& Relevant evidence & She puts any unused items in a giant fire outside her cave. \\
& Distractor evidence & The Coldoc cleans everything in her cave three times a day. \\
& Correct answer & No, because it might get burned. \\
& Distractor answer & She probably won’t notice, so go ahead. \\
& Distractor answer & No, it might get lost in her messy cave. \\
\midrule

\multirow{6}{*}{3}
& Question & Which feeling did the Anpert have at school? \\
& Relevant evidence & The Anpert ran to the cafeteria and piled her plate with four pizza slices and a salad. \\
& Distractor evidence & The Anpert skipped and danced as she walked out of the principal’s office. \\
& Correct answer & Hungry \\
& Distractor answer & Angry \\
& Distractor answer & Tired \\
\bottomrule
\end{tabular*}

\vspace{0.5em}
\begin{minipage}{\textwidth}
\footnotesize
\textit{Note:} Each item requires students to select both a correct answer and the supporting evidence sentence.
\end{minipage}
\end{table}

\newpage

\section{Appendix C. Distribution of Text Information}\label{AppendixC}
\begin{figure}
    \centering
    \includegraphics[width=.8\linewidth]{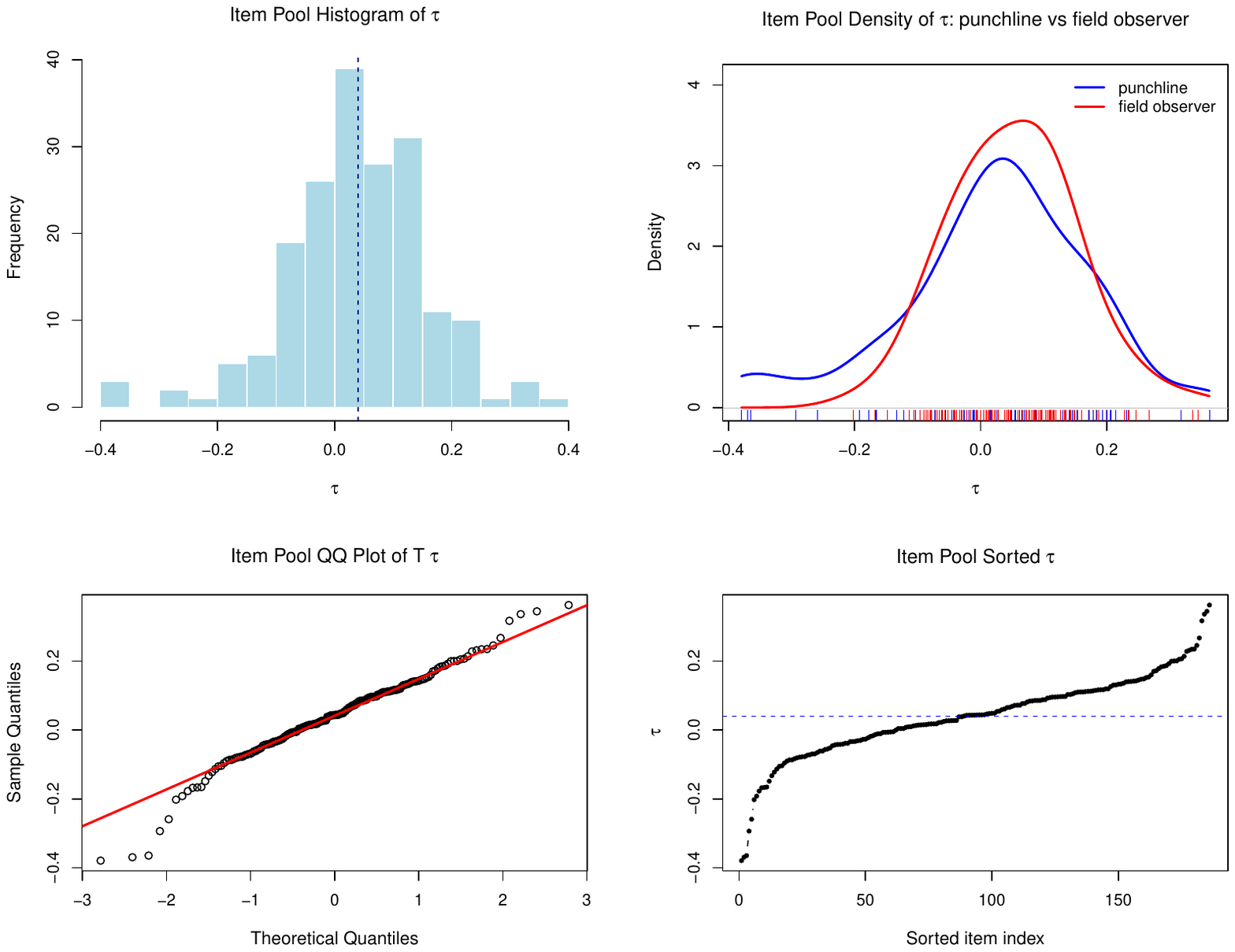}
    \caption{Distribution of $\tau$ values in the item pool}
    \label{fig:Tdistri}
\end{figure}

\begin{figure}
    \centering
    \includegraphics[width=.8\linewidth]{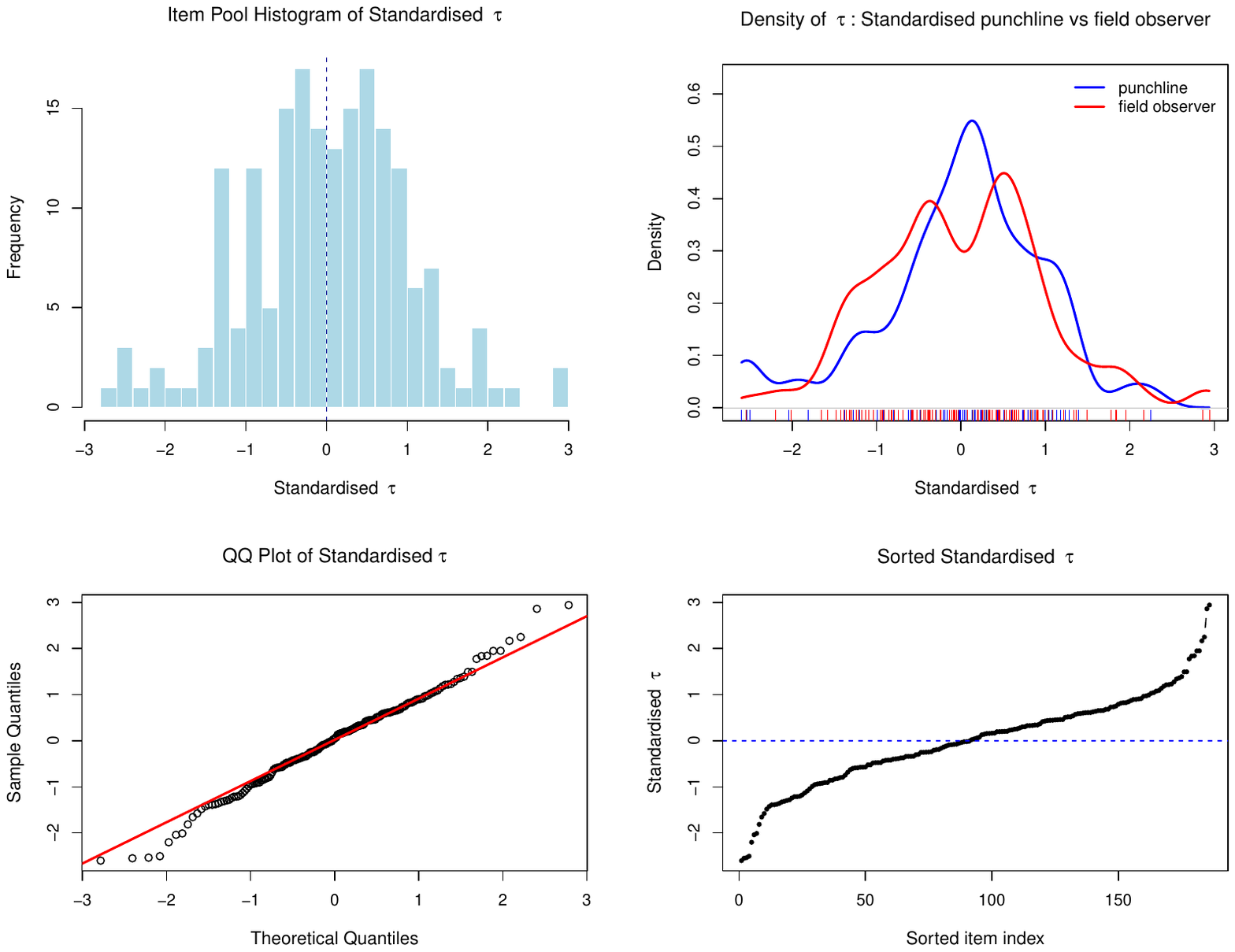}
    \caption{Distribution of standardised $\tau$ values in the item pool}
    \label{fig:Tdistri_standard}
\end{figure}

The distribution of the item pool is shown in Figure \ref{fig:Tdistri}, and the standardized version is shown in Figure \ref{fig:Tdistri_standard}. Because both Grade 2 and Grade 3 were drawn from the same underlying item pool, the distribution of item-level $\tau$ is identical across grades. Differences between grades arise only from the item sampling process and subsequent student responses. The histogram and QQ plot indicate that $\tau$ is approximately normally distributed, with only mild deviations in the tails. The density panel shows that the two games differ in shape: \textit{Punchline}
items form a single peak near zero, whereas \textit{Field Observer} items are
bimodal, with one mode just below zero and a second above it. The sorted values reveal a smooth distribution, providing no evidence of extreme outliers.

\section{Appendix D. True \texorpdfstring{$Q$}{Q}-matrix in Simulation}

The shorter test forms were constructed as nested subsets of the full 24-item test. The true $Q$-matrices for the simulation with $J = 24$ and $K = 2$ are shown in Table~\ref{tab:qmatrix_j24_t1_t2}. The $Q$-matrices were held fixed across all simulation replicates.

\begin{table}[H]
\centering
\small
\caption{True $Q$-matrices for $J = 24$ at Time 1 and Time 2}
\label{tab:qmatrix_j24_t1_t2}
\setlength{\tabcolsep}{4pt}
\begin{tabular}{c|cc|cc@{\qquad}c|cc|cc}
\toprule
\multirow{2}{*}{Item} &
\multicolumn{2}{c|}{\textbf{Time 1}} &
\multicolumn{2}{c}{\textbf{Time 2}} &
\multirow{2}{*}{Item} &
\multicolumn{2}{c|}{\textbf{Time 1}} &
\multicolumn{2}{c}{\textbf{Time 2}} \\
\cmidrule(lr){2-3} \cmidrule(lr){4-5}
\cmidrule(lr){7-8} \cmidrule(lr){9-10}
& $A_1$ & $A_2$ & $A_1$ & $A_2$
& & $A_1$ & $A_2$ & $A_1$ & $A_2$ \\
\midrule
1  & 1 & 0 & 1 & 0 & 13 & 1 & 0 & 1 & 1 \\
2  & 1 & 0 & 0 & 1 & 14 & 0 & 1 & 1 & 1 \\
3  & 0 & 1 & 1 & 0 & 15 & 1 & 1 & 1 & 1 \\
4  & 0 & 1 & 0 & 1 & 16 & 1 & 0 & 1 & 1 \\
5  & 1 & 0 & 1 & 1 & 17 & 1 & 1 & 1 & 1 \\
6  & 1 & 1 & 1 & 0 & 18 & 1 & 1 & 0 & 1 \\
7  & 0 & 1 & 1 & 0 & 19 & 1 & 1 & 1 & 1 \\
8  & 0 & 1 & 1 & 1 & 20 & 0 & 1 & 0 & 1 \\
9  & 1 & 0 & 1 & 1 & 21 & 0 & 1 & 1 & 1 \\
10 & 0 & 1 & 1 & 1 & 22 & 1 & 0 & 1 & 0 \\
11 & 1 & 0 & 1 & 1 & 23 & 0 & 1 & 0 & 1 \\
12 & 1 & 0 & 0 & 1 & 24 & 0 & 1 & 1 & 1 \\
\bottomrule
\end{tabular}
\end{table}

\end{document}

%% file: empirical_tables/table_emp_q_item_joint_stepwise.tex
\begin{table}[ht]
\centering
\caption{Text-prior empirical estimates of the $Q$-matrix, guessing ($g$), and slipping ($s$) parameters from the joint and stepwise approaches. Stepwise estimates are from the first-step measurement models fitted separately at each time point.}
\label{tab:emp_q_item_joint_stepwise}
\scriptsize
\resizebox{\textwidth}{!}{
\begin{tabular}{c|cccc|cccc|cccc|cccc}
\toprule
\multirow{3}{*}{Item} & \multicolumn{8}{c|}{\textbf{Time 1}} & \multicolumn{8}{c}{\textbf{Time 2}} \\
\cmidrule(lr){2-9} \cmidrule(lr){10-17}
& \multicolumn{4}{c|}{\textbf{Joint}} & \multicolumn{4}{c|}{\textbf{Stepwise}} & \multicolumn{4}{c|}{\textbf{Joint}} & \multicolumn{4}{c}{\textbf{Stepwise}} \\
\cmidrule(lr){2-5} \cmidrule(lr){6-9} \cmidrule(lr){10-13} \cmidrule(lr){14-17}
& $A_1$ & $A_2$ & $g$ & $s$ & $A_1$ & $A_2$ & $g$ & $s$ & $A_1$ & $A_2$ & $g$ & $s$ & $A_1$ & $A_2$ & $g$ & $s$ \\
\midrule
1 & 0 & 1 & 0.191 & 0.045 & 1 & 1 & 0.232 & 0.032 & 1 & 1 & 0.586 & 0.030 & 1 & 1 & 0.543 & 0.022 \\
2 & 0 & 1 & 0.208 & 0.019 & 1 & 1 & 0.252 & 0.013 & 0 & 1 & 0.557 & 0.026 & 1 & 1 & 0.615 & 0.023 \\
3 & 0 & 1 & 0.252 & 0.084 & 1 & 1 & 0.288 & 0.072 & 0 & 1 & 0.732 & 0.046 & 1 & 0 & 0.706 & 0.047 \\
4 & 1 & 0 & 0.172 & 0.034 & 0 & 1 & 0.141 & 0.054 & 1 & 1 & 0.753 & 0.029 & 1 & 1 & 0.756 & 0.033 \\
5 & 1 & 0 & 0.258 & 0.026 & 0 & 1 & 0.229 & 0.042 & 1 & 0 & 0.865 & 0.012 & 1 & 1 & 0.885 & 0.013 \\
6 & 1 & 0 & 0.342 & 0.013 & 0 & 1 & 0.325 & 0.036 & 1 & 1 & 0.867 & 0.010 & 1 & 1 & 0.866 & 0.011 \\
7 & 0 & 1 & 0.954 & 0.024 & 1 & 1 & 0.953 & 0.020 & 0 & 1 & 0.640 & 0.025 & 0 & 1 & 0.554 & 0.030 \\
8 & 1 & 0 & 0.914 & 0.048 & 0 & 1 & 0.911 & 0.045 & 0 & 1 & 0.825 & 0.014 & 0 & 1 & 0.777 & 0.015 \\
9 & 1 & 1 & 0.850 & 0.089 & 1 & 1 & 0.849 & 0.083 & 0 & 1 & 0.580 & 0.042 & 0 & 1 & 0.483 & 0.047 \\
10 & 1 & 0 & 0.909 & 0.043 & 1 & 1 & 0.911 & 0.039 & 0 & 1 & 0.708 & 0.039 & 0 & 1 & 0.678 & 0.045 \\
11 & 1 & 0 & 0.904 & 0.048 & 0 & 1 & 0.903 & 0.049 & 0 & 1 & 0.848 & 0.010 & 0 & 1 & 0.818 & 0.010 \\
12 & 1 & 0 & 0.936 & 0.039 & 1 & 1 & 0.937 & 0.039 & 0 & 1 & 0.613 & 0.063 & 1 & 1 & 0.670 & 0.064 \\
\bottomrule
\end{tabular}}
\end{table}

%% file: empirical_tables/table_emp_profile_transition_joint_stepwise.tex
\begin{table}[ht]
\centering
\caption{Text-prior transition matrices of attribute profiles from Time 1 to Time 2 under the joint and stepwise approaches. Each cell reports count (percentage). Stepwise counts are based on first-step hard classifications before the third-step CEP correction is applied to structural estimation. Profile labels: 00 = no mastery, 10 = $A_1$ only, 01 = $A_2$ only, and 11 = mastery of both attributes.}
\label{tab:emp_profile_transition_joint_stepwise}
\setlength{\tabcolsep}{3pt}
\scriptsize
\begin{tabular}{llcccc|c}
\toprule
\textbf{Method} & \textbf{Time 1} & \multicolumn{4}{c|}{\textbf{Time 2}} & \textbf{Totals} \\
\cmidrule(lr){3-6}
& & 00 & 10 & 01 & 11 & \\
\midrule
\multirow{5}{*}{Joint}
& 00 & 68 (3.44\%) & 42 (2.12\%) & 39 (1.97\%) & 537 (27.15\%) & 686 (34.68\%) \\
& 10 & 13 (0.66\%) & 39 (1.97\%) & 4 (0.20\%) & 439 (22.19\%) & 495 (25.03\%) \\
& 01 & 13 (0.66\%) & 5 (0.25\%) & 11 (0.56\%) & 114 (5.76\%) & 143 (7.23\%) \\
& 11 & 1 (0.05\%) & 11 (0.56\%) & 2 (0.10\%) & 640 (32.36\%) & 654 (33.06\%) \\
& \textbf{Totals} & 95 (4.80\%) & 97 (4.90\%) & 56 (2.83\%) & 1730 (87.46\%) & 1978 (100.00\%) \\
\midrule
\multirow{5}{*}{Stepwise}
& 00 & 10 (0.51\%) & 40 (2.02\%) & 101 (5.11\%) & 430 (21.74\%) & 581 (29.37\%) \\
& 10 & 1 (0.05\%) & 13 (0.66\%) & 27 (1.37\%) & 135 (6.83\%) & 176 (8.90\%) \\
& 01 & 3 (0.15\%) & 22 (1.11\%) & 58 (2.93\%) & 527 (26.64\%) & 610 (30.84\%) \\
& 11 & 3 (0.15\%) & 6 (0.30\%) & 35 (1.77\%) & 567 (28.67\%) & 611 (30.89\%) \\
& \textbf{Totals} & 17 (0.86\%) & 81 (4.10\%) & 221 (11.17\%) & 1659 (83.87\%) & 1978 (100.00\%) \\
\bottomrule
\end{tabular}
\end{table}

%% file: empirical_tables/table_emp_significant_covariates_joint_stepwise.tex
\begin{table}[htbp]
\centering
\caption{Significant covariates for initial mastery ($\beta_z$) and acquisition transition probability ($\gamma_{01}$) under the text-prior joint and stepwise approaches. Only covariates with 95\% credible intervals for odds ratios excluding 1 are shown. \textit{Note}. OR = odds ratio; RT = response time; NLM = number of correctly answered questions; NRA = number of attempts; WB = well below benchmark.}
\label{tab:emp_significant_covariates_joint_stepwise}
\scriptsize
\begin{tabular}{ccc|ccc}
\toprule
\multicolumn{3}{c|}{\textbf{Initial mastery}} & \multicolumn{3}{c}{\textbf{Initial mastery}} \\
\multicolumn{3}{c|}{\textbf{Joint}} & \multicolumn{3}{c}{\textbf{Stepwise}} \\
$K$ & Covariate & OR & $K$ & Covariate & OR \\
\midrule
1 & RT Punchline & \textbf{1.319} & 1 & NLM Punchline & \textbf{11.918} \\
1 & NLM Field Observer & \textbf{1.198} & 1 & NRA Punchline & \textbf{0.069} \\
1 & NLM Punchline & \textbf{9.392} & 2 & RT Punchline & \textbf{1.613} \\
1 & NRA Punchline & \textbf{0.297} & 2 & NLM Field Observer & \textbf{1.290} \\
1 & Benchmark-WB & \textbf{0.465} & 2 & NLM Punchline & \textbf{9.045} \\
2 & NLM Punchline & \textbf{8.440} & 2 & NRA Punchline & \textbf{0.270} \\
2 & NRA Punchline & \textbf{0.094} & 2 & Benchmark-WB & \textbf{0.445} \\
\midrule
\multicolumn{3}{c|}{\textbf{Transition probability}} & \multicolumn{3}{c}{\textbf{Transition probability}} \\
\multicolumn{3}{c|}{\textbf{Joint}} & \multicolumn{3}{c}{\textbf{Stepwise}} \\
$K$ & Covariate & OR & $K$ & Covariate & OR \\
\midrule
1 & NLM Field Observer & \textbf{1.361} & 1 & NLM Field Observer & \textbf{1.330} \\
1 & Benchmark-WB & \textbf{0.404} & 1 & Benchmark-WB & \textbf{0.292} \\
2 & RT Punchline & \textbf{1.412} & 2 & NLM Field Observer & \textbf{1.424} \\
2 & NLM Field Observer & \textbf{1.456} & 2 & NLM Punchline & \textbf{0.673} \\
2 & Benchmark-WB & \textbf{0.382} & 2 & NRA Field Observer & \textbf{0.691} \\
 & & & 2 & Benchmark-AB & \textbf{2.723} \\
\bottomrule
\end{tabular}
\end{table}

%% file: empirical_tables/table_emp_betaZ_OR_part1_joint_stepwise.tex
\begin{sidewaystable}[htbp]
\centering
\caption{Posterior means of odds ratios (OR) for initial mastery covariate effects $\beta_z$ under the text-prior joint and stepwise approaches, with 95\% credible intervals. Statistically significant results are shown in \textbf{bold}. Part 1 of 2.\\\textit{Note}: RT = response time; NLM = number of correctly answered questions; NRA = number of attempts.}
\label{tab:emp_betaZ_OR_part1_joint_stepwise}
\begin{tabularx}{\linewidth}{lll*{7}{>{\centering\arraybackslash}X}}
\toprule
\textbf{Method} & \textbf{K} & \textbf{Measure} & RT Field Observer & RT Punchline & NLM Field Observer & NLM Punchline & NRA Field Observer & NRA Punchline & Gender (male) \\
\midrule
Joint & 1 & OR & 1.029 & \textbf{1.319} & \textbf{1.198} & \textbf{9.392} & 0.925 & \textbf{0.297} & 1.809 \\
 &  & CI & (0.892, 1.168) & (1.063, 1.719) & (1.054, 1.384) & (7.464, 12.206) & (0.755, 1.129) & (0.213, 0.385) & (0.159, 7.620) \\
 & 2 & OR & 1.035 & 0.955 & 1.098 & \textbf{8.440} & 0.868 & \textbf{0.094} & 1.717 \\
 &  & CI & (0.881, 1.193) & (0.803, 1.124) & (0.878, 1.339) & (6.365, 11.490) & (0.666, 1.120) & (0.063, 0.136) & (0.129, 8.131) \\
\addlinespace
Stepwise & 1 & OR & 1.004 & 0.928 & 1.340 & \textbf{11.918} & 0.956 & \textbf{0.069} & 1.447 \\
 &  & CI & (0.797, 1.214) & (0.681, 1.234) & (0.977, 1.750) & (7.609, 18.095) & (0.634, 1.392) & (0.037, 0.112) & (0.179, 5.563) \\
 & 2 & OR & 1.002 & \textbf{1.613} & \textbf{1.290} & \textbf{9.045} & 0.924 & \textbf{0.270} & 1.835 \\
 &  & CI & (0.865, 1.161) & (1.203, 2.138) & (1.103, 1.510) & (6.580, 11.937) & (0.737, 1.126) & (0.192, 0.360) & (0.127, 9.043) \\
\bottomrule
\end{tabularx}
\end{sidewaystable}

%% file: empirical_tables/table_emp_betaZ_OR_part2_joint_stepwise.tex
\begin{sidewaystable}[htbp]
\centering
\caption{Posterior means of odds ratios (OR) for initial mastery covariate effects $\beta_z$ under the text-prior joint and stepwise approaches, with 95\% credible intervals. Statistically significant results are shown in \textbf{bold}. Part 2 of 2.\\\textit{Note}: SEN = special educational needs; ELL = English language learner; WB = well below benchmark; BB = below benchmark; AB = above benchmark. Race effects are relative to White students.}
\label{tab:emp_betaZ_OR_part2_joint_stepwise}
\begin{tabularx}{\linewidth}{lll*{7}{>{\centering\arraybackslash}X}}
\toprule
\textbf{Method} & \textbf{K} & \textbf{Measure} & SEN & ELL & Benchmark-WB & Benchmark-BB & Benchmark-AB & Race (Asian) & Race (URM) \\
\midrule
Joint & 1 & OR & 1.650 & 1.521 & \textbf{0.465} & 0.723 & 1.134 & 1.622 & 0.903 \\
 &  & CI & (0.111, 6.292) & (0.132, 6.748) & (0.209, 0.929) & (0.350, 1.386) & (0.760, 1.634) & (0.161, 7.227) & (0.214, 2.960) \\
 & 2 & OR & 1.544 & 2.041 & 0.531 & 0.578 & 0.763 & 1.697 & 0.499 \\
 &  & CI & (0.126, 6.244) & (0.159, 8.871) & (0.211, 1.116) & (0.224, 1.153) & (0.507, 1.160) & (0.141, 7.013) & (0.165, 1.143) \\
\addlinespace
Stepwise & 1 & OR & 1.566 & 1.564 & 0.493 & 0.516 & 0.707 & 1.539 & 0.646 \\
 &  & CI & (0.143, 6.974) & (0.145, 6.899) & (0.138, 1.144) & (0.190, 1.101) & (0.379, 1.220) & (0.165, 6.436) & (0.151, 1.641) \\
 & 2 & OR & 1.693 & 1.557 & \textbf{0.445} & 0.689 & 1.019 & 1.624 & 2.762 \\
 &  & CI & (0.127, 7.302) & (0.118, 7.694) & (0.180, 0.924) & (0.292, 1.417) & (0.711, 1.513) & (0.133, 7.010) & (0.366, 10.585) \\
\bottomrule
\end{tabularx}
\end{sidewaystable}

%% file: empirical_tables/table_emp_gamma01_OR_part1_joint_stepwise.tex
\begin{sidewaystable}[htbp]
\centering
\caption{Posterior means of odds ratios (OR) for acquisition transition effects $\gamma_{01}$ under the text-prior joint and stepwise approaches, with 95\% credible intervals. Statistically significant results are shown in \textbf{bold}. Part 1 of 2.\\\textit{Note}: RT = response time; NLM = number of correctly answered questions; NRA = number of attempts.}
\label{tab:emp_gamma01_OR_part1_joint_stepwise}
\begin{tabularx}{\linewidth}{lll*{7}{>{\centering\arraybackslash}X}}
\toprule
\textbf{Method} & \textbf{K} & \textbf{Measure} & RT Field Observer & RT Punchline & NLM Field Observer & NLM Punchline & NRA Field Observer & NRA Punchline & Gender (male) \\
\midrule
Joint & 1 & OR & 0.979 & 1.312 & \textbf{1.361} & 0.968 & 0.878 & 1.157 & 1.581 \\
 &  & CI & (0.622, 1.448) & (0.795, 2.283) & (1.068, 1.731) & (0.663, 1.326) & (0.659, 1.235) & (0.870, 1.527) & (0.122, 7.113) \\
 & 2 & OR & 1.105 & \textbf{1.412} & \textbf{1.456} & 0.990 & 0.843 & 1.011 & 1.670 \\
 &  & CI & (0.851, 1.551) & (1.027, 2.043) & (1.194, 1.829) & (0.618, 1.437) & (0.627, 1.091) & (0.837, 1.257) & (0.136, 6.487) \\
\addlinespace
Stepwise & 1 & OR & 0.988 & 1.306 & \textbf{1.330} & 1.252 & 0.870 & 1.112 & 1.409 \\
 &  & CI & (0.754, 1.321) & (0.921, 2.028) & (1.089, 1.627) & (0.918, 1.679) & (0.683, 1.122) & (0.845, 1.524) & (0.153, 5.267) \\
 & 2 & OR & 1.149 & 1.495 & \textbf{1.424} & \textbf{0.673} & \textbf{0.691} & 1.111 & 1.528 \\
 &  & CI & (0.710, 2.190) & (0.763, 2.686) & (1.056, 1.883) & (0.415, 0.958) & (0.517, 0.871) & (0.857, 1.442) & (0.125, 6.247) \\
\bottomrule
\end{tabularx}
\end{sidewaystable}

%% file: empirical_tables/table_emp_gamma01_OR_part2_joint_stepwise.tex
\begin{sidewaystable}[htbp]
\centering
\caption{Posterior means of odds ratios (OR) for acquisition transition effects $\gamma_{01}$ under the text-prior joint and stepwise approaches, with 95\% credible intervals. Statistically significant results are shown in \textbf{bold}. Part 2 of 2.\\\textit{Note}: SEN = special educational needs; ELL = English language learner; WB = well below benchmark; BB = below benchmark; AB = above benchmark. Race effects are relative to White students.}
\label{tab:emp_gamma01_OR_part2_joint_stepwise}
\begin{tabularx}{\linewidth}{lll*{7}{>{\centering\arraybackslash}X}}
\toprule
\textbf{Method} & \textbf{K} & \textbf{Measure} & SEN & ELL & Benchmark-WB & Benchmark-BB & Benchmark-AB & Race (Asian) & Race (URM) \\
\midrule
Joint & 1 & OR & 1.774 & 1.642 & \textbf{0.404} & 0.953 & 2.542 & 1.449 & 1.284 \\
 &  & CI & (0.149, 7.525) & (0.160, 6.814) & (0.110, 0.963) & (0.324, 2.458) & (0.824, 5.485) & (0.167, 4.919) & (0.365, 3.836) \\
 & 2 & OR & 1.702 & 1.648 & \textbf{0.382} & 0.635 & 1.502 & 1.577 & 2.572 \\
 &  & CI & (0.142, 8.193) & (0.162, 6.763) & (0.138, 0.783) & (0.223, 1.534) & (0.490, 3.365) & (0.127, 7.236) & (0.352, 6.300) \\
\addlinespace
Stepwise & 1 & OR & 1.595 & 1.702 & \textbf{0.292} & 0.470 & 0.633 & 1.509 & 3.511 \\
 &  & CI & (0.136, 6.300) & (0.143, 7.961) & (0.087, 0.763) & (0.168, 1.010) & (0.358, 1.052) & (0.118, 5.979) & (0.634, 10.474) \\
 & 2 & OR & 1.831 & 1.784 & 0.431 & 0.942 & \textbf{2.723} & 1.681 & 2.608 \\
 &  & CI & (0.157, 8.089) & (0.148, 7.815) & (0.104, 1.061) & (0.257, 2.463) & (1.252, 5.279) & (0.130, 7.755) & (0.526, 8.219) \\
\bottomrule
\end{tabularx}
\end{sidewaystable}

%% file: simu_tables/table_Q_recovery_joint_stepwise.tex
\begin{landscape}
\begin{table}[!htbp]
\centering
\caption{Recovery of the $Q$-matrix under the joint and stepwise approaches. Values are reported as mean (bootstrap SE)}
\label{tab:q_recovery_joint_stepwise}
\footnotesize
\setlength{\tabcolsep}{1.8pt}
\begin{threeparttable}
\begin{tabular}{ccc|ccccc|ccccc}
\toprule
\multicolumn{3}{c|}{} & \multicolumn{5}{c|}{Joint} & \multicolumn{5}{c}{Stepwise} \\
\cmidrule(lr){4-8} \cmidrule(lr){9-13}
$N$ & $J_t$ & $T$ & ACC & FPR & FNR & PIP RMSE & PIP MAE & ACC & FPR & FNR & PIP RMSE & PIP MAE \\
\midrule

1000 & 6 & 1 
& 1.000 (0.000) & 0.000 (0.000) & 0.000 (0.000) & 0.000 (0.000) & 0.000 (0.000) 
& 1.000 (0.000) & 0.000 (0.000) & 0.000 (0.000) & 0.014 (0.006) & 0.005 (0.002) \\

& & 2 
& 1.000 (0.000) & 0.000 (0.000) & 0.000 (0.000) & 0.000 (0.000) & 0.000 (0.000) 
& 0.750 (0.064) & 0.327 (0.077) & 0.195 (0.056) & 0.362 (0.044) & 0.327 (0.041) \\

\addlinespace[0.3em]

2000 & 6 & 1 
& 1.000 (0.000) & 0.000 (0.000) & 0.000 (0.000) & 0.000 (0.000) & 0.000 (0.000) 
& 1.000 (0.000) & 0.000 (0.000) & 0.000 (0.000) & 0.009 (0.006) & 0.003 (0.002) \\

& & 2 
& 1.000 (0.000) & 0.000 (0.000) & 0.000 (0.000) & 0.000 (0.000) & 0.000 (0.000) 
& 0.593 (0.080) & 0.500 (0.098) & 0.341 (0.064) & 0.444 (0.042) & 0.404 (0.039) \\

\addlinespace[0.3em]

4000 & 6 & 1 
& 1.000 (0.000) & 0.000 (0.000) & 0.000 (0.000) & 0.000 (0.000) & 0.000 (0.000) 
& 1.000 (0.000) & 0.000 (0.000) & 0.000 (0.000) & 0.008 (0.008) & 0.003 (0.003) \\

& & 2 
& 1.000 (0.000) & 0.000 (0.000) & 0.000 (0.000) & 0.000 (0.000) & 0.000 (0.000) 
& 0.608 (0.100) & 0.494 (0.122) & 0.319 (0.084) & 0.491 (0.065) & 0.439 (0.064) \\

\addlinespace[0.3em]

1000 & 12 & 1 
& 1.000 (0.000) & 0.000 (0.000) & 0.000 (0.000) & 0.030 (0.029) & 0.028 (0.027) 
& 0.993 (0.007) & 0.017 (0.016) & 0.000 (0.000) & 0.071 (0.023) & 0.057 (0.019) \\

& & 2 
& 1.000 (0.000) & 0.000 (0.000) & 0.000 (0.000) & 0.024 (0.022) & 0.017 (0.016) 
& 0.714 (0.043) & 0.589 (0.087) & 0.185 (0.029) & 0.369 (0.039) & 0.257 (0.028) \\

\addlinespace[0.3em]

2000 & 12 & 1 
& 1.000 (0.000) & 0.000 (0.000) & 0.000 (0.000) & 0.030 (0.029) & 0.028 (0.026) 
& 1.000 (0.000) & 0.000 (0.000) & 0.000 (0.000) & 0.041 (0.022) & 0.038 (0.020) \\

& & 2 
& 1.000 (0.000) & 0.000 (0.000) & 0.000 (0.000) & 0.000 (0.000) & 0.000 (0.000) 
& 0.843 (0.041) & 0.402 (0.100) & 0.076 (0.025) & 0.291 (0.035) & 0.203 (0.025) \\

\addlinespace[0.3em]

4000 & 12 & 1 
& 1.000 (0.000) & 0.000 (0.000) & 0.000 (0.000) & 0.000 (0.000) & 0.000 (0.000) 
& 1.000 (0.000) & 0.000 (0.000) & 0.000 (0.000) & 0.121 (0.054) & 0.103 (0.046) \\

& & 2 
& 1.000 (0.000) & 0.000 (0.000) & 0.000 (0.000) & 0.000 (0.000) & 0.000 (0.000) 
& 0.714 (0.095) & 0.571 (0.187) & 0.190 (0.062) & 0.377 (0.078) & 0.266 (0.057) \\

\addlinespace[0.3em]

1000 & 24 & 1 
& 1.000 (0.000) & 0.000 (0.000) & 0.000 (0.000) & 0.029 (0.028) & 0.025 (0.024) 
& 0.917 (0.047) & 0.111 (0.060) & 0.067 (0.036) & 0.180 (0.037) & 0.155 (0.032) \\

& & 2 
& 1.000 (0.000) & 0.000 (0.000) & 0.000 (0.000) & 0.000 (0.000) & 0.000 (0.000) 
& 0.738 (0.039) & 0.630 (0.093) & 0.166 (0.025) & 0.351 (0.038) & 0.226 (0.023) \\

\addlinespace[0.3em]

2000 & 24 & 1 
& 1.000 (0.000) & 0.000 (0.000) & 0.000 (0.000) & 0.024 (0.022) & 0.021 (0.019) 
& 0.932 (0.067) & 0.091 (0.088) & 0.055 (0.052) & 0.131 (0.055) & 0.114 (0.049) \\

& & 2 
& 1.000 (0.000) & 0.000 (0.000) & 0.000 (0.000) & 0.000 (0.000) & 0.000 (0.000) 
& 0.744 (0.057) & 0.636 (0.143) & 0.156 (0.037) & 0.368 (0.038) & 0.237 (0.026) \\

\addlinespace[0.3em]

4000 & 24 & 1 
& 1.000 (0.000) & 0.000 (0.000) & 0.000 (0.000) & 0.000 (0.000) & 0.000 (0.000) 
& 0.965 (0.040) & 0.045 (0.038) & 0.028 (0.024) & 0.095 (0.048) & 0.081 (0.041) \\

& & 2 
& 1.000 (0.000) & 0.000 (0.000) & 0.000 (0.000) & 0.000 (0.000) & 0.000 (0.000) 
& 0.820 (0.072) & 0.420 (0.160) & 0.110 (0.041) & 0.255 (0.055) & 0.171 (0.038) \\

\bottomrule
\end{tabular}

\begin{tablenotes}[flushleft]
\footnotesize
\item[] \textit{Note.} PIP RMSE and PIP MAE summarize posterior inclusion probability accuracy. 
\end{tablenotes}

\end{threeparttable}
\end{table}
\end{landscape}

%% file: simu_tables/table_alpha_recovery_joint_stepwise.tex
\begin{table}[!htbp]
\centering
\caption{Recovery of attribute profiles under the joint and stepwise approaches. Values are reported as mean (bootstrap SE)}
\label{tab:alpha_recovery_joint_stepwise}
\footnotesize
\setlength{\tabcolsep}{3.5pt}
\begin{threeparttable}
\begin{tabular}{ccc|ccc|ccc}
\toprule
\multicolumn{3}{c|}{} & \multicolumn{3}{c|}{Joint} & \multicolumn{3}{c}{Stepwise} \\
\cmidrule(lr){4-6} \cmidrule(lr){7-9}
$N$ & $J_t$ & $T$ & PAR & AAR$_1$ & AAR$_2$ & PAR & AAR$_1$ & AAR$_2$ \\
\midrule
1000 & 6 & 1 & 0.925 (0.006) & 0.976 (0.003) & 0.949 (0.004) & 0.891 (0.004) & 0.958 (0.002) & 0.930 (0.004) \\
 &  & 2 & 0.918 (0.005) & 0.974 (0.002) & 0.942 (0.005) & 0.750 (0.029) & 0.850 (0.032) & 0.785 (0.029) \\
\addlinespace[0.3em]
2000 & 6 & 1 & 0.934 (0.003) & 0.975 (0.002) & 0.958 (0.003) & 0.899 (0.003) & 0.960 (0.002) & 0.935 (0.003) \\
 &  & 2 & 0.918 (0.005) & 0.975 (0.003) & 0.942 (0.004) & 0.687 (0.040) & 0.772 (0.038) & 0.727 (0.038) \\
\addlinespace[0.3em]
4000 & 6 & 1 & 0.937 (0.005) & 0.976 (0.004) & 0.960 (0.002) & 0.899 (0.005) & 0.960 (0.003) & 0.935 (0.004) \\
 &  & 2 & 0.930 (0.004) & 0.969 (0.003) & 0.959 (0.002) & 0.693 (0.047) & 0.767 (0.051) & 0.739 (0.047) \\
\addlinespace[0.3em]
1000 & 12 & 1 & 0.976 (0.001) & 0.996 (0.001) & 0.979 (0.001) & 0.959 (0.008) & 0.992 (0.001) & 0.967 (0.008) \\
 &  & 2 & 0.943 (0.006) & 0.992 (0.001) & 0.950 (0.007)  & 0.673 (0.036) & 0.735 (0.038) & 0.699 (0.037) \\
\addlinespace[0.3em]
2000 & 12 & 1 & 0.971 (0.002) & 0.994 (0.001) & 0.976 (0.003) & 0.969 (0.002) & 0.992 (0.001) & 0.977 (0.002) \\
 &  & 2 & 0.938 (0.006) & 0.988 (0.002) & 0.948 (0.007) & 0.779 (0.040) & 0.872 (0.030) & 0.809 (0.038) \\
\addlinespace[0.3em]
4000 & 12 & 1 & 0.978 (0.001) & 0.995 (0.001) & 0.982 (0.001) & 0.963 (0.003) & 0.992 (0.001) & 0.971 (0.003) \\
 &  & 2 & 0.951 (0.003) & 0.989 (0.001) & 0.960 (0.003) & 0.684 (0.081) & 0.735 (0.080) & 0.704 (0.079) \\
\addlinespace[0.3em]
1000 & 24 & 1 & 0.989 (0.001) & 0.998 (0.001) & 0.991 (0.001) & 0.945 (0.025) & 0.952 (0.025) & 0.951 (0.022) \\
 &  & 2 & 0.954 (0.015) & 0.996 (0.000) & 0.958 (0.015) & 0.668 (0.038) & 0.701 (0.044) & 0.717 (0.030) \\
\addlinespace[0.3em]
2000 & 24 & 1 & 0.981 (0.002) & 0.999 (0.000) & 0.983 (0.002) & 0.946 (0.034) & 0.963 (0.034) & 0.950 (0.033) \\
 &  & 2 & 0.946 (0.007) & 0.995 (0.001) & 0.951 (0.007) & 0.669 (0.059) & 0.711 (0.065) & 0.700 (0.053) \\
\addlinespace[0.3em]
4000 & 24 & 1 & 0.974 (0.007) & 0.999 (0.000) & 0.975 (0.007) & 0.950 (0.033) & 0.967 (0.032) & 0.954 (0.031) \\
 &  & 2 & 0.950 (0.007) & 0.995 (0.001) & 0.955 (0.008) & 0.670 (0.055) & 0.715 (0.060) & 0.705 (0.052) \\
\bottomrule
\end{tabular}
\begin{tablenotes}[flushleft]
\footnotesize
\item[] \textit{Note.} PAR denotes profile agreement rate; AAR denotes attribute-wise agreement rate. 
\end{tablenotes}
\end{threeparttable}
\end{table}

%% file: simu_tables/table_item_recovery_joint_stepwise.tex
\begin{landscape}
\begin{table}[p]
\centering
\vspace*{\fill}

\caption{Estimation accuracy of item parameters under the joint and stepwise approaches. Values are reported as mean (bootstrap SE)}
\label{tab:item_recovery_joint_stepwise}

\renewcommand{\arraystretch}{1.08}
\setlength{\tabcolsep}{2.5pt}

\begin{threeparttable}
\begin{adjustbox}{max width=1\linewidth,center}
\begin{tabular}{ccc|cccc|cccc}
\toprule
\multicolumn{3}{c|}{} & \multicolumn{4}{c|}{Joint} & \multicolumn{4}{c}{Stepwise} \\
\cmidrule(lr){4-7} \cmidrule(lr){8-11}
$N$ & $J_t$ & Metric & $g_{t1}$ & $g_{t2}$ & $s_{t1}$ & $s_{t2}$ & $g_{t1}$ & $g_{t2}$ & $s_{t1}$ & $s_{t2}$ \\
\midrule
1000 & 6 & RMSE & 0.013 (0.001) & 0.019 (0.002) & 0.028 (0.003) & 0.015 (0.001) & 0.032 (0.008) & 0.041 (0.008) & 0.052 (0.008) & 0.032 (0.006) \\
 &  & MAE & 0.011 (0.001) & 0.016 (0.001) & 0.022 (0.003) & 0.013 (0.001) & 0.023 (0.005) & 0.030 (0.005) & 0.041 (0.006) & 0.025 (0.004) \\
\addlinespace[0.3em]
2000 & 6 & RMSE & 0.009 (0.000) & 0.013 (0.001) & 0.021 (0.002) & 0.011 (0.001) & 0.024 (0.008) & 0.028 (0.008) & 0.042 (0.010) & 0.021 (0.005) \\
 &  & MAE  & 0.008 (0.000) & 0.009 (0.001) & 0.018 (0.002) & 0.009 (0.000) & 0.019 (0.006) & 0.020 (0.005) & 0.034 (0.009) & 0.016 (0.003) \\
\addlinespace[0.3em]
4000 & 6 & RMSE & 0.007 (0.001) & 0.009 (0.001) & 0.015 (0.002) & 0.008 (0.001) & 0.009 (0.002) & 0.035 (0.012) & 0.022 (0.006) & 0.026 (0.007) \\
 &  & MAE & 0.005 (0.001) & 0.007 (0.001) & 0.013 (0.002) & 0.007 (0.001) & 0.007 (0.001) & 0.023 (0.007) & 0.017 (0.004) & 0.020 (0.005) \\
\addlinespace[0.3em]
1000 & 12 & RMSE & 0.013 (0.001) & 0.022 (0.002) & 0.023 (0.002) & 0.015 (0.001) & 0.014 (0.001) & 0.022 (0.003) & 0.039 (0.006) & 0.017 (0.001) \\
 &  & MAE & 0.010 (0.001) & 0.016 (0.001) & 0.019 (0.001) & 0.012 (0.001) & 0.011 (0.001) & 0.017 (0.002) & 0.031 (0.005) & 0.014 (0.001) \\
\addlinespace[0.3em]
2000 & 12 & RMSE & 0.010 (0.002) & 0.020 (0.005) & 0.018 (0.001) & 0.013 (0.001) & 0.009 (0.000) & 0.033 (0.004) & 0.020 (0.001) & 0.014 (0.001) \\
 &  & MAE & 0.008 (0.001) & 0.013 (0.003) & 0.014 (0.001) & 0.010 (0.001) & 0.007 (0.000) & 0.021 (0.002) & 0.015 (0.001) & 0.011 (0.001) \\
\addlinespace[0.3em]
4000 & 12 & RMSE & 0.007 (0.000) & 0.013 (0.002) & 0.016 (0.001) & 0.009 (0.001) & 0.012 (0.003) & 0.021 (0.008) & 0.046 (0.016) & 0.013 (0.002) \\
 &  & MAE & 0.006 (0.000) & 0.009 (0.001) & 0.012 (0.001) & 0.007 (0.001) & 0.009 (0.002) & 0.014 (0.005) & 0.033 (0.012) & 0.009 (0.001) \\
\addlinespace[0.3em]
1000 & 24 & RMSE & 0.014 (0.002) & 0.029 (0.006) & 0.024 (0.002) & 0.016 (0.002) & 0.015 (0.001) & 0.031 (0.002) & 0.026 (0.001) & 0.016 (0.000) \\
 &  & MAE & 0.011 (0.002) & 0.020 (0.003) & 0.019 (0.001) & 0.012 (0.001) & 0.012 (0.001) & 0.021 (0.001) & 0.020 (0.001) & 0.013 (0.000) \\
\addlinespace[0.3em]
2000 & 24 & RMSE & 0.016 (0.002) & 0.034 (0.003) & 0.019 (0.001) & 0.012 (0.001) & 0.010 (0.001) & 0.023 (0.003) & 0.019 (0.001) & 0.013 (0.001) \\
 &  & MAE & 0.013 (0.002) & 0.021 (0.002) & 0.015 (0.001) & 0.010 (0.001) & 0.008 (0.001) & 0.016 (0.002) & 0.014 (0.001) & 0.010 (0.000) \\
\addlinespace[0.3em]
4000 & 24 & RMSE & 0.019 (0.004) & 0.032 (0.004) & 0.014 (0.000) & 0.010 (0.001) & 0.008 (0.001) & 0.019 (0.003) & 0.016 (0.001) & 0.011 (0.001) \\
 &  & MAE & 0.014 (0.003) & 0.018 (0.002) & 0.011 (0.000) & 0.008 (0.001) & 0.006 (0.001) & 0.013 (0.002) & 0.012 (0.001) & 0.009 (0.001) \\
\bottomrule
\end{tabular}
\end{adjustbox}

\begin{tablenotes}[flushleft]
\footnotesize
\item[] \textit{Note.} $g$ and $s$ denote guessing and slipping parameters at Time 1 and Time 2.
\end{tablenotes}
\end{threeparttable}

\vspace*{\fill}
\end{table}
\end{landscape}

%% file: simu_tables/table_regression_recovery_joint_stepwise.tex
\begin{table}[!htbp]
\centering
\caption{Estimation accuracy of regression parameters under the joint and stepwise approaches. Values are reported as mean (bootstrap SE).}
\label{tab:regression_recovery_joint_stepwise}
\footnotesize
\setlength{\tabcolsep}{3pt}
\begin{threeparttable}
\begin{tabular}{ccc|ccc|ccc}
\toprule
\multicolumn{3}{c|}{} & \multicolumn{3}{c|}{Joint} & \multicolumn{3}{c}{Stepwise} \\
\cmidrule(lr){4-6} \cmidrule(lr){7-9}
$N$ & $J_t$ & Metric & $\beta_0$ & $\beta_Z$ & $\gamma_{01}$ & $\beta_0$ & $\beta_Z$ & $\gamma_{01}$ \\
\midrule
1000 & 6 & RMSE & 0.088 (0.014) & 0.096 (0.008) & 0.095 (0.006) & 0.202 (0.035) & 0.127 (0.014) & 0.414 (0.033) \\
 &  & MAE & 0.077 (0.013) & 0.080 (0.007) & 0.076 (0.006) & 0.167 (0.026) & 0.101 (0.010) & 0.322 (0.025) \\
\addlinespace[0.3em]
2000 & 6 & RMSE & 0.087 (0.014) & 0.062 (0.004) & 0.072 (0.003) & 0.099 (0.009) & 0.115 (0.030) & 0.395 (0.035) \\
 &  & MAE & 0.080 (0.013) & 0.051 (0.003) & 0.058 (0.002) & 0.092 (0.009) & 0.082 (0.015) & 0.316 (0.027) \\
\addlinespace[0.3em]
4000 & 6 & RMSE & 0.070 (0.017) & 0.044 (0.003) & 0.052 (0.003) & 0.080 (0.028) & 0.049 (0.001) & 0.362 (0.037) \\
 &  & MAE & 0.062 (0.016) & 0.035 (0.003) & 0.045 (0.003) & 0.066 (0.021) & 0.039 (0.001) & 0.285 (0.031) \\
\addlinespace[0.3em]
1000 & 12 & RMSE & 0.089 (0.020) & 0.096 (0.015) & 0.116 (0.005) & 0.168 (0.030) & 0.139 (0.025) & 0.431 (0.033) \\
 &  & MAE & 0.075 (0.017) & 0.077 (0.012) & 0.091 (0.006) & 0.148 (0.025) & 0.106 (0.018) & 0.339 (0.026) \\
\addlinespace[0.3em]
2000 & 12 & RMSE & 0.090 (0.018) & 0.057 (0.005) & 0.080 (0.013) & 0.081 (0.009) & 0.072 (0.007) & 0.326 (0.033) \\
 &  & MAE & 0.075 (0.015) & 0.047 (0.004) & 0.060 (0.007) & 0.071 (0.007) & 0.058 (0.005) & 0.246 (0.023) \\
\addlinespace[0.3em]
4000 & 12 & RMSE & 0.061 (0.010) & 0.042 (0.003) & 0.050 (0.006) & 0.217 (0.073) & 0.132 (0.047) & 0.409 (0.068) \\
 &  & MAE & 0.052 (0.008) & 0.034 (0.003) & 0.041 (0.004) & 0.186 (0.061) & 0.101 (0.034) & 0.324 (0.058) \\
\addlinespace[0.3em]
1000 & 24 & RMSE & 0.100 (0.029) & 0.073 (0.004) & 0.117 (0.019) & 0.189 (0.030) & 0.262 (0.050) & 0.408 (0.020) \\
 &  & MAE & 0.088 (0.025) & 0.056 (0.004) & 0.083 (0.010) & 0.169 (0.026) & 0.201 (0.037) & 0.322 (0.017) \\
\addlinespace[0.3em]
2000 & 24 & RMSE & 0.140 (0.027) & 0.084 (0.011) & 0.099 (0.009) & 0.154 (0.045) & 0.211 (0.075) & 0.428 (0.029) \\
 &  & MAE & 0.121 (0.024) & 0.069 (0.009) & 0.075 (0.006) & 0.140 (0.039) & 0.157 (0.059) & 0.333 (0.024) \\
\addlinespace[0.3em]
4000 & 24 & RMSE & 0.148 (0.040) & 0.054 (0.005) & 0.074 (0.007) & 0.165 (0.050) & 0.190 (0.065) & 0.390 (0.035) \\
 &  & MAE & 0.120 (0.028) & 0.042 (0.002) & 0.054 (0.007) & 0.145 (0.042) & 0.140 (0.050) & 0.305 (0.028) \\
\bottomrule
\end{tabular}
\begin{tablenotes}[flushleft]
\footnotesize
\item[] \textit{Note.} $\beta_0$ denotes initial mastery intercepts, $\beta_Z$ denotes initial mastery covariate effects, and $\gamma_{01}$ denotes acquisition transition parameters. 
\end{tablenotes}
\end{threeparttable}
\end{table}